%% file: main.tex
\documentclass{article}

\usepackage{iclr2027_conference,times}
\input{math_commands.tex}

\usepackage{amsmath}
\usepackage{amssymb}
\usepackage{booktabs}
\usepackage{graphicx}
\usepackage{microtype}
\usepackage{multicol}
\usepackage{multirow}
\usepackage{needspace}
\usepackage[table]{xcolor}
\usepackage{xspace}
\usepackage{hyperref}
\usepackage{url}
\hypersetup{hidelinks}

\makeatletter
\renewcommand{\section}{\@startsection{section}{1}{\z@}{-1.6ex plus -0.4ex minus -0.2ex}{0.8ex plus 0.2ex minus 0.1ex}{\large\sc\raggedright}}
\renewcommand{\subsection}{\@startsection{subsection}{2}{\z@}{-1.35ex plus -0.4ex minus -0.2ex}{0.45ex plus 0.15ex}{\normalsize\sc\raggedright}}
\makeatother

\title{DepthEvidence: Unifying Metric Depth Prediction and Geometric Reasoning in Multimodal Language Models}

\author{
\textbf{Jiangning Wei\textsuperscript{1,2}\quad
Yuan Yao\textsuperscript{1}\quad
Miaomiao Cui\textsuperscript{1}\quad
Mingsheng Li\textsuperscript{1} \quad
Humen Zhong\textsuperscript{1}\quad} \\
\textbf{Shuai Bai\textsuperscript{1}\quad
Zhibo Yang\textsuperscript{1}}\\
\textsuperscript{1}Alibaba Token Hub, Alibaba Group \\
\textsuperscript{2}Beijing University of Posts and Telecommunications
}

\newcommand{\authorversion}{}
\ifdefined\authorversion
\iclrfinalcopy
\fi

\begin{document}

\maketitle
\ifdefined\authorversion
\fancyhead[L]{}
\fi

\begin{abstract}
\input{sections/00_abstract}
\end{abstract}

\addtocontents{toc}{\protect\noindent\protect\textsc{Main Paper}\protect\par\protect\smallskip}
\input{sections/01_introduction}
\input{sections/02_related_work}
\input{sections/04_method}
\input{sections/05_instance_depth_benchmark}
\input{sections/05_experimental_setup}
\input{sections/06_main_results}
\input{sections/07_analysis}
\input{sections/08_conclusion}

\newpage

\subsection*{AI Use Statement}

We used Qwen3.7-plus and Seed-2.1-Turbo to propose object annotations and to generate and cross-verify synthetic Depth-VQA questions and answers, as described in Section 4 and Appendix C.1. We additionally used generative AI tools to improve manuscript readability. All benchmark examples were manually verified, and the authors reviewed all AI-assisted outputs and take responsibility for the final content.

\subsection*{Reproducibility Statement}
Section~\ref{sec:method} and Appendix~\ref{app:method_details} describe the model architecture and training objectives, while Section~\ref{sec:experimental_setup} and Appendix~\ref{app:training_details} provide the training pipeline, data composition, preprocessing, and optimization settings. Section~\ref{sec:instance_depth_benchmark} and Appendix~\ref{app:evaluation_protocols} document benchmark construction, data partitioning, evaluation protocols, and metric definitions.

\phantomsection
\addcontentsline{toc}{section}{References}
\bibliography{references}
\bibliographystyle{iclr2027_conference}

\clearpage
\addtocontents{toc}{\protect\columnbreak\protect\noindent\protect\textsc{Appendix}\protect\par\protect\smallskip}
\appendix
\begingroup
\setcounter{tocdepth}{2}
\footnotesize
\setlength{\columnsep}{2em}
\makeatletter
\renewcommand{\l@section}[2]{%
  \addvspace{0.25em}%
  \@dottedtocline{1}{0em}{1.5em}{\bfseries #1}{\bfseries #2}}
\renewcommand{\l@subsection}[2]{%
  \@dottedtocline{2}{1.5em}{2.5em}{#1}{#2}}
\section*{Paper and Appendix Contents}
\vspace{-0.4ex}
\begin{multicols}{2}
\raggedcolumns
\@starttoc{toc}
\end{multicols}
\makeatother
\endgroup
\input{sections/09_appendix}

\end{document}

%% file: math_commands.tex
\usepackage{amsmath,amsfonts,bm}

\def\eqref#1{equation~\ref{#1}}

\def\1{\bm{1}}

\DeclareMathAlphabet{\mathsfit}{\encodingdefault}{\sfdefault}{m}{sl}
\SetMathAlphabet{\mathsfit}{bold}{\encodingdefault}{\sfdefault}{bx}{n}



%% file: sections/00_abstract.tex
Spatial reasoning with metric constraints requires linking objects to geometric measurements and preserving their numerical content during language reasoning. We present DepthEvidence, a 4B model that uses its own dense metric predictions as object-grounded evidence for language generation. A camera-conditioned decoder predicts full-resolution metric depth using multi-scale visual features and high-resolution RGB refinement. A dense-to-language interface converts predicted depths and decoder features into object-aligned continuous geometry tokens anchored to object identifiers. Geometric supervision encourages metric information to remain recoverable before and after language-context interaction, while instruction tuning supports object measurement and compositional reasoning. We introduce a Depth-VQA benchmark evaluating object-depth queries, relative comparisons, and decisions combining spatial and numerical constraints. Across nine datasets, DepthEvidence achieves the highest average dense $\delta_1$ among evaluated methods, competitive with specialized estimators. It also leads the evaluated methods in instance-level metric depth estimation and overall accuracy on both relative and metric reasoning tracks, while broadly preserving general VQA performance and improving spatial understanding relative to the base model.

%% file: sections/01_introduction.tex
\section{Introduction}
\label{sec:introduction}

\begin{figure}[th]
    \centering
    \includegraphics[width=1.0\linewidth]{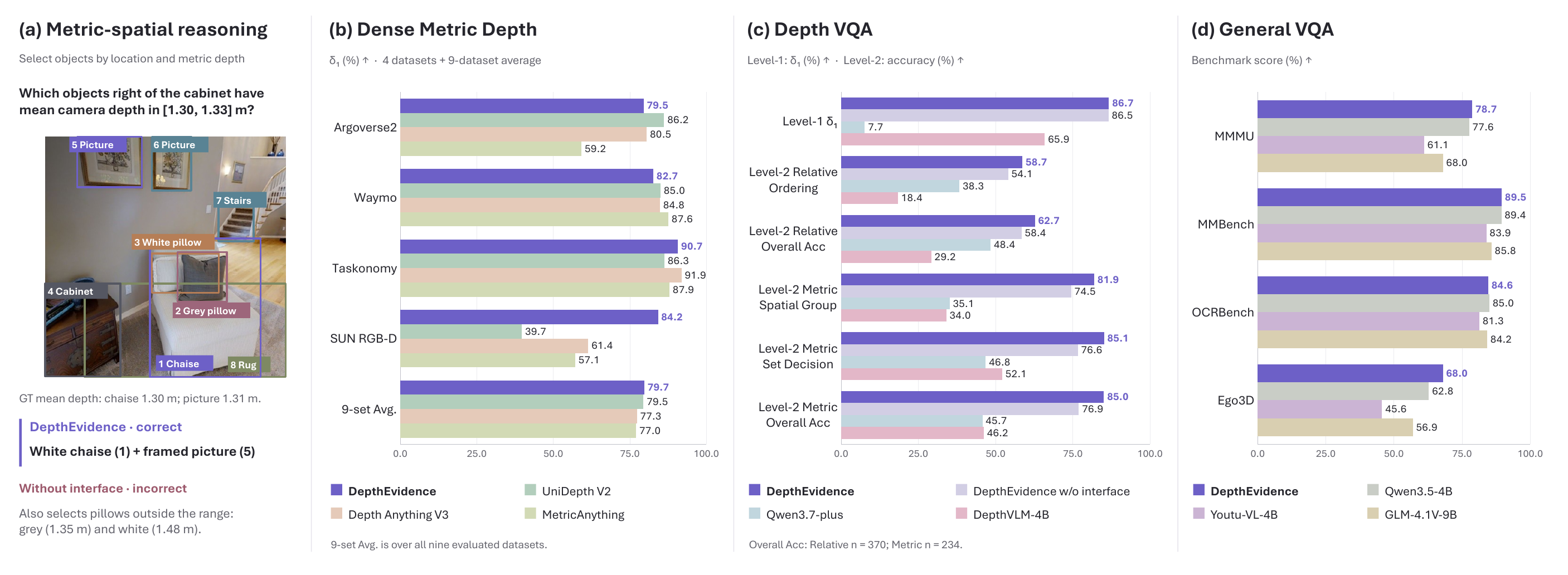}
    \caption{Overview of DepthEvidence's capabilities. (a) A metric-spatial question requires selecting objects by image location and mean camera depth; the dense-to-language interface rejects out-of-range distractors. (b) Dense metric depth results ($\delta_1$) on four representative datasets and the nine-dataset average, compared with specialized estimators. (c) Level~1 metric estimation and Level~2 relative and metric reasoning, compared with general-purpose and depth-aware VLMs as well as the no-interface variant. (d) General VQA retention on MMMU, MMBench, and OCRBench, and spatial understanding on Ego3D. All scores are percentages and higher is better.}
    \label{fig:capability_overview}
\end{figure}

Answering spatial questions with metric constraints requires connecting object semantics to geometric measurements. Knowing which object is closer establishes a relative relation, but selecting objects within a depth interval requires absolute scale and the correct association between each object and its measured depth. Consider the question in Figure~\ref{fig:capability_overview}(a): \emph{which referenced objects to the right of a wooden cabinet have an average camera depth between 1.30\,m and 1.33\,m?} The answer must satisfy both the spatial filter and the numerical interval. Dense metric estimators provide pixel-level measurements~\citep{yin2023metric3d,piccinelli2024unidepth}, while spatially aware multimodal large language models (MLLMs) connect objects with language~\citep{chen2024spatialvlm,cheng2024spatialrgpt}. This question requires bringing those capabilities together at the level of individual objects.

Recent depth-aware VLMs have substantially advanced geometric prediction. DepthLM~\citep{cai2026depthlm} demonstrates accurate pixel-level metric estimation through language modeling, using visual prompting and intrinsic-conditioned augmentation without a dedicated depth head. DepthVLM~\citep{yu2026depthvlm} extends this capability to full-resolution dense prediction through a lightweight head and joint depth--text training, while retaining general multimodal capability and improving spatial reasoning. G$^2$VLM also couples geometric reconstruction with spatial reasoning~\citep{hu2026g2vlm}. Building on these advances, we ask: \emph{how can a VLM use its own dense metric predictions as object-grounded evidence for language reasoning?} We focus on explicitly linking the predicted metric field to the referenced objects and preserving its numerical content as the language model processes the question.

Using predicted geometry as language evidence requires both object alignment and metric information preservation. Measurements must come from the appropriate image regions and remain associated with the correct references as the model processes multiple objects. Their numerical content must also survive projection into language tokens and interaction with the question. Answer supervision targets the final response without directly constraining the metric content of intermediate geometry tokens. This motivates explicit geometric supervision to keep object-level depth information recoverable before and after interaction with the language context.

We present DepthEvidence, a 4B model that turns its own dense predictions into object-aligned evidence for language generation. Given an image and camera intrinsics, a camera-conditioned decoder combines multi-scale visual features with high-resolution RGB refinement to produce metric depth and preserve local structure. The dense-to-language interface pools predicted depths and decoder features within each referenced region, then encodes them as object-aligned continuous geometry tokens anchored to object identifiers. Auxiliary depth objectives supervise metric information in these tokens before and after language-context interaction, while a ranking objective supervises object ordering. The tokens enter answer generation through cached continuation, avoiding repeated encoding of the image and question. In the cabinet example, this interface supplies the candidate objects' geometry for spatial filtering and metric-range selection.

To evaluate how geometric prediction translates into language answers, we organize depth understanding into dense metric depth prediction (Level~0), instance-level metric depth estimation (Level~1), and compositional depth reasoning (Level~2). Instance-level instruction tuning and the interface support both measurement and reasoning. Our Depth-VQA benchmark evaluates whether models can turn object-level depth measurements into compositional answers, covering relative comparisons and decisions that jointly depend on spatial relations and metric gaps, ranges, or thresholds. DepthEvidence achieves the highest average dense $\delta_1$ across nine datasets among the evaluated methods and the highest overall accuracy on both reasoning tracks, while general VQA performance remains broadly stable relative to the base model (Figure~\ref{fig:capability_overview}).

Our contributions are:
\begin{itemize}
    \item An object-aligned dense-to-language interface with geometric supervision before and after contextualization, enabling language generation to condition on the model's own metric predictions.
    \item DepthEvidence, a unified 4B model combining camera-conditioned dense prediction with instance-level depth instruction tuning, together with a Depth-VQA benchmark that distinguishes object measurement, relative reasoning, and metric-constrained reasoning.
    \item DepthEvidence leads the evaluated methods in nine-dataset average dense $\delta_1$, instance-level metric estimation, and overall compositional reasoning, while retaining general VQA performance and improving spatial understanding over the base model.
\end{itemize}

%% file: sections/02_related_work.tex
\section{Related Work}
\label{sec:related_work}

\paragraph{Monocular Metric Depth Estimation.}
Monocular depth estimation has progressed from multi-scale convolutional models with scale-invariant supervision~\citep{eigen2014depth} to transformer decoders using multi-level visual tokens~\citep{ranftl2021vision}. Depth Anything and Depth Anything V2 scale transferable representations with unlabeled or pseudo-labeled data~\citep{yang2024depthanything,yang2024depthanythingv2}, whereas Metric3D and UniDepth address camera variation through canonicalization or camera-conditioned features~\citep{yin2023metric3d,piccinelli2024unidepth}. These methods primarily treat geometry as a dense output; DepthEvidence additionally uses metric predictions as object-grounded evidence for measurement and reasoning.

\paragraph{Geometry-Aware Vision--Language Models.}
Prior work uses geometry either to support spatial reasoning or as an explicit model output. SpatialVLM constructs metric spatial question--answer data through automatic 3D annotation~\citep{chen2024spatialvlm}, and SpatialRGPT combines 3D scene-graph regions with a depth-aware visual plugin~\citep{cheng2024spatialrgpt}; G$^2$VLM, Vid-LLM, and SpatialStack more directly couple reconstruction, 3D tokens, or geometry-encoder features with language reasoning~\citep{hu2026g2vlm,chen2026vidllm,zhang2026spatialstack}. AURORA and Perceptio emit discrete depth-related tokens before text~\citep{bigverdi2025perceptiontokens,li2026perceptio}, DepthLM predicts metric depth at prompted pixels~\citep{cai2026depthlm}, and DepthVLM jointly outputs full-resolution depth and text~\citep{yu2026depthvlm}. However, explicit or parallel depth prediction does not necessarily condition answers on the predicted metric field. DepthEvidence bridges prediction and reasoning by converting its single-image dense outputs into object-aligned continuous geometry tokens, supervised before and after language-context interaction, for object-level measurement and relative or metric compositional reasoning.

%% file: sections/04_method.tex
\section{Method}
\label{sec:method}

\subsection{Framework Overview}

\begin{figure}[t!]
    \centering
    \includegraphics[width=0.75 \linewidth ]{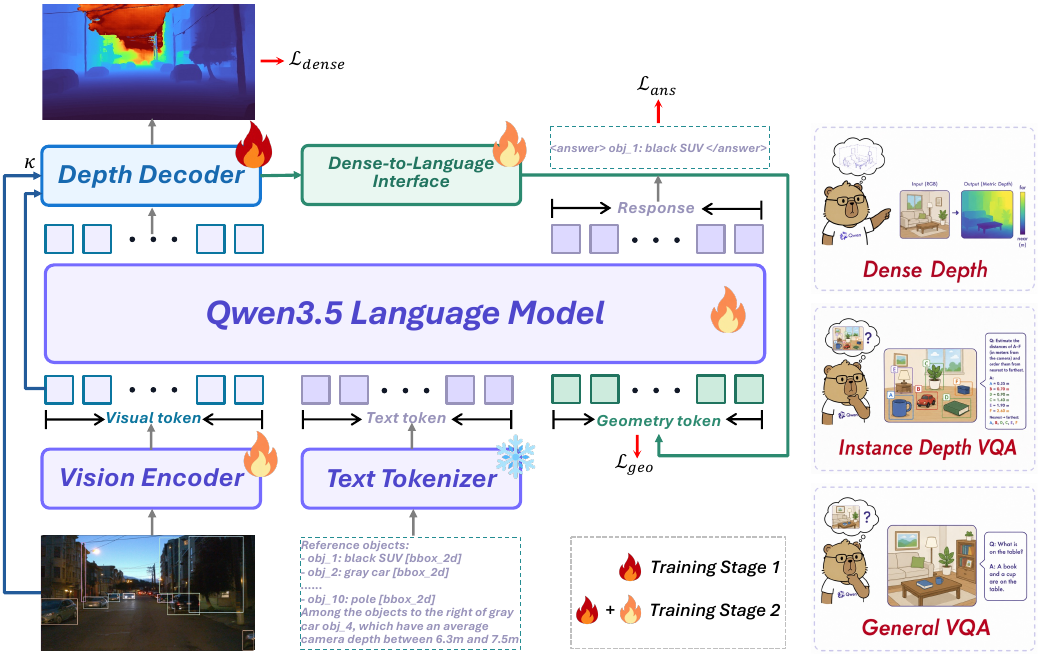}
    \caption{Overview of DepthEvidence. The dense decoder supplies metric predictions and features to the dense-to-language interface, which converts them into object-aligned geometry tokens for answer generation. This path connects Level-0 dense prediction to Level-1 instance measurement and Level-2 compositional reasoning within a shared VLM.}
    \label{fig:overall_architecture}
\end{figure}

As illustrated in Figure~\ref{fig:overall_architecture}, DepthEvidence conditions object-level answers on geometric evidence predicted from the same image. A camera-conditioned dense decoder supplies a metric depth map and spatial features. The dense-to-language interface extracts evidence for objects specified by boxes or referring prompts and binds their geometry tokens to explicit object identifiers. Auxiliary objectives supervise metric-depth prediction from these tokens before and after language-context interaction. Instance-level instruction tuning trains the VLM to use this context for measurement and compositional reasoning.

\subsection{Level-0 Dense Metric Depth Prediction}
\label{sec:dense_depth}

\paragraph{Camera-conditioned decoder.}
The Level-0 branch follows a coarse-to-fine DPT design~\citep{ranftl2021vision}, fusing intermediate vision-encoder features with the spatially reshaped final visual hidden states. To recover metric scale, it converts the supplied camera intrinsics $\boldsymbol{\kappa}$ into a compact descriptor containing normalized calibration, field-of-view, and aspect-ratio information. An MLP maps this descriptor to FiLM parameters~\citep{perez2018film} that condition every decoder scale. A lightweight high-resolution RGB pathway supplies finer appearance cues during upsampling, and the resulting full-resolution representation predicts both the metric depth map and the dense features used by the dense-to-language interface. This numerical camera encoding is specific to the dense branch; Depth-VQA instead receives the raw intrinsic values in its instruction. Appendix~\ref{app:dense_decoder} provides the feature dimensions, camera parameterization, and decoder details.

\paragraph{High-resolution RGB refinement.}
Spatially compressed visual tokens can lose thin structures and sharp depth discontinuities. The RGB pathway therefore injects high-resolution features through learned residual gates, preserving the transformer representation as the main path while restoring local boundary cues. The refined spatial features also provide local appearance and boundary cues for object-level pooling in the interface. Architectural details are given in Appendix~\ref{app:rgb_refinement}.

\paragraph{Training objective.}
We train the dense branch with
\begin{equation}\mathcal{L}_{\mathrm{dense}}=0.8\mathcal{L}_{\mathrm{abs}}+0.2\mathcal{L}_{\mathrm{silog}}+0.1\mathcal{L}_{\mathrm{der}}+0.03\mathcal{L}_{\mathrm{lap}}+\mathcal{L}_{\mathrm{ssi}}.\label{eq:dense_training_objective}\end{equation}
The first two terms supervise metric depth, the derivative and Laplacian terms preserve multi-scale structure, and the edge-guided stable local scale-and-shift invariant term $\mathcal{L}_{\mathrm{ssi}}$ emphasizes object boundaries. Appendix~\ref{app:dense_objective} provides the detailed definitions and implementation.

\subsection{Instance-Level Depth Instruction Tuning}
\label{sec:instance_depth_tuning}

\paragraph{Instance-level measurement and reasoning tasks.}
We extend dense prediction with two object-grounded instruction levels. Level~1 asks for a specified object's center, mean, median, nearest, or farthest metric depth. Level~2 combines measurements across objects: relative tasks cover near--far and front--behind comparisons, closest/farthest selection, and subset or full depth ordering; metric tasks cover numerical mean-depth gaps, thresholds, ranges, and margin-based decisions. Compositional questions first resolve image-space references or filter candidates, then apply relative comparisons or metric constraints. These instructions teach object-level measurement and its use in spatially and numerically constrained answers.

\paragraph{Unified object grounding.}
We support two ways to specify an object. A box prompt provides its category and bounding box, with coordinates normalized to a $0$--$1000$ image coordinate system. A referring prompt instead identifies it through a natural-language description enclosed by object-reference tokens. For multi-object questions, each instance receives a stable identifier such as \texttt{obj\_1}, and the prompt lists its identifier, category, and grounding cue before the question. This common serialization supports both grounding modes and disambiguates repeated object categories without assuming any interface-specific processing.

\paragraph{Structured supervision.}
Each instruction supplies the raw camera intrinsics as \texttt{<focal> $f_x$ $f_y$ $c_x$ $c_y$ </focal>}, independently of the dense decoder's numerical camera encoding. The VLM reads this block together with the image and question. We train both instruction levels with the next-token cross-entropy loss $\mathcal{L}_{\mathrm{ans}}$ on the final answer. The answer is a metric depth value for Level~1, or a numerical depth gap, an object or set selection, an ordering, or a discrete comparison or constraint decision for Level~2. Margin-based decisions include an \emph{ambiguous} answer when the closest object does not satisfy the required depth margin.
Appendix~\ref{app:instance_instruction} provides the detailed targets, prompt serialization, and output formats.

\subsection{Dense-to-Language Interface}
\label{sec:dense_to_language}

\paragraph{Object-aligned dense geometry.}
For both Level~1 and Level~2 instructions, the interface receives a metric depth map $\widehat{\mathbf{D}}$ and a full-resolution feature map $\mathbf{F}^{d}$ from the dense decoder. For a box prompt, we use the supplied region directly; for a referring prompt, a lightweight grounding head predicts its box. For object $i$, pooling $\mathbf{F}^{d}$ yields a regional feature $\mathbf{r}_i$, while summarizing the predicted depths yields metric statistics $\mathbf{s}_i$. Together with the normalized box $\mathbf{b}_i$, these form an object descriptor that associates local geometry with the referenced instance. Appendix~\ref{app:geometry_interface_details} gives the pooling statistics and grounding-head implementation.

\paragraph{Continuous geometry tokens.}
A two-layer projector $P_g$ maps each object descriptor to $K=2$ continuous tokens,
\begin{equation}\mathbf{g}_{i,k}=P_g([\mathbf{r}_i;\mathbf{b}_i;\mathbf{s}_i])_k+\mathbf{p}_i+\mathbf{t}_k,\qquad k\in\{1,2\},\label{eq:geometry_tokens}\end{equation}
where $\mathbf{p}_i$ is a learned object-index embedding and $\mathbf{t}_k$ distinguishes the two token slots. We organize the resulting tokens as an input-only geometry bank in which each pair follows its textual anchor \texttt{obj\_i:}. Geometry-bank positions are excluded as language-modeling targets, but answer tokens attend to them and the answer loss backpropagates through them to the projector. Appendix~\ref{app:geometry_interface_details} details the bank construction, and Appendix~\ref{app:two_geometry_tokens} discusses the choice of $K=2$.

\paragraph{Geometric supervision.}
To encourage metric-depth recoverability, we supervise the geometry representations at both instance levels with
\begin{equation}\mathcal{L}_{\mathrm{geo}}=0.2\mathcal{L}_{\mathrm{raw}}+0.2\mathcal{L}_{\mathrm{ctx}}+0.1\mathcal{L}_{\mathrm{rank}},\label{eq:geometry_training_objective}\end{equation}
The raw-depth and contextual-depth losses train auxiliary heads to predict five ground-truth object-depth statistics (center, mean, median, nearest, and farthest) in log space, from the projected tokens and their contextualized states, respectively. These objectives encourage metric-depth recoverability before and after interaction with the image and question. For multi-object examples, $\mathcal{L}_{\mathrm{rank}}$ supervises relative ordering using the contextual mean-depth predictions. Appendix~\ref{app:geometry_objectives} specifies the heads, targets, and gradient routing.

\paragraph{Single-prefill cached continuation.}
The geometry bank becomes available after the image and question have been encoded. We retain the prefix cache, append the bank as a causal continuation, and generate the answer from the extended context. Geometry tokens can attend to the cached prefix, and answer tokens can attend to both the prefix and geometry. This execution scheme incorporates predicted evidence without re-encoding the multimodal prefix. Appendix~\ref{app:cached_geometry_details} gives implementation details, and Appendix~\ref{app:inference_efficiency} compares no-interface, two-pass, and single-prefill inference.

%% file: sections/05_instance_depth_benchmark.tex
\section{Instance-Level Depth-VQA Benchmark}
\label{sec:instance_depth_benchmark}

Depth-VQA evaluates object-level metric estimation (Level~1) and multi-object compositional reasoning (Level~2), including relative comparisons, ordering, numerical gaps, and spatially conditioned metric decisions. Each question provides an image, camera intrinsics, and object references specified by normalized category--box prompts or natural-language referring prompts.

\paragraph{Construction and quality control.}
We first complete sparse outdoor depth so every source image has a dense depth map. Qwen3.7-plus~\citep{qwen2026qwen37plus} and Seed-2.1-Turbo~\citep{bytedance2026seed21} independently identify major objects and propose their categories and grounding boxes, reducing dependence on the output distribution of a single generator. We pool their candidates, remove boxes covering less than 1\% of the image or lying near image boundaries, prompt SAM~3~\citep{carion2025sam3} with the remaining categories and boxes, and apply a second quality filter to the resulting masks. Each retained mask and dense depth map yield the Level~1 center, mean, median, nearest, and farthest targets, with the latter two defined by the 5th and 95th depth percentiles. The two generators then produce Level~2 questions and answers from these instance records, with each candidate verified by the model that did not generate it. We manually audit 1\% of training samples across all three levels to derive targeted filters, manually check every benchmark example, and limit reuse of the same instance across task types within a video scene. Appendix~\ref{app:benchmark_construction} provides full details.

\paragraph{Level-1: Instance-Level Metric Depth Estimation.}

Level~1 contains 757 grounded instances from six indoor and outdoor datasets: Waymo~\citep{sun2020waymo}, Argoverse~2~\citep{wilson2021argoverse2}, Taskonomy~\citep{zamir2018taskonomy}, LingBot-Depth~\citep{tan2026lingbotdepth}, DDAD~\citep{guizilini2020ddad}, and nuScenes~\citep{caesar2020nuscenes}. Each query requests the object-center, nearest, farthest, mean, or median valid depth within the referenced mask and expects a scalar answer in meters. We report $\delta_1$ (higher is better), RMSE, and AbsRel (lower is better).

\paragraph{Level-2: Compositional Depth Reasoning.}

Depth-VQA contains 1,503 questions in total: 757 Level~1 questions, 370 Level~2 Relative questions, and 376 Level~2 Metric questions. Level~2 uses grounded instances as reasoning operands, assigning each a stable identifier such as \texttt{obj\_1}. Table~\ref{tab:level2_benchmark} groups seven subbenches into the two tracks; Appendix Table~\ref{tab:level2_benchmark_full} defines all question types.
The Relative track uses exact-match accuracy: selections require the target identifier, and orderings require the full identifier sequence. Relative Overall Acc pools the 370 questions from the three reported subbenches. The Metric track uses accuracy for discrete answers and $\delta_1$ for numerical gaps; Metric Overall Acc excludes $\delta_1$-scored questions. Appendix~\ref{app:evaluation_protocols} provides question counts and scoring details.

\begin{table}[t]
    \centering
    \caption{Level-2 tracks and subbench goals. Appendix Table~\ref{tab:level2_benchmark_full} provides all question types and reasoning requirements.}
    \label{tab:level2_benchmark}
    \scriptsize
    \setlength{\tabcolsep}{4pt}
    \renewcommand{\arraystretch}{0.9}
    \begin{minipage}{0.92\linewidth}
    \centering
    \begin{tabular}{@{}p{0.14\linewidth}p{0.20\linewidth}p{0.61\linewidth}@{}}
        \toprule
        Track & Subbench & Question summary \\
        \midrule
        \multirow{3}{=}{Relative Depth} & Spatial Pair & Compare two objects, optionally after resolving a spatial reference. \\
        & Spatial Group & Spatially filter candidates, then select the nearest or farthest. \\
        & Depth Ordering & Select an extreme or order a subset or all referenced objects. \\
        \midrule
        \multirow{4}{=}{Metric Depth} & Metric Pair & Predict a mean-depth gap or compare it with a threshold. \\
        & Spatial Metric Pair & Resolve a spatial reference, then predict an absolute mean-depth gap. \\
        & Spatial Metric Group & Spatially filter candidates, then apply range or bound constraints. \\
        & Metric Set Decision & Test a safety band or a margin-constrained closest-object decision. \\
        \bottomrule
    \end{tabular}
    \end{minipage}
\end{table}

%% file: sections/05_experimental_setup.tex
\section{Experimental Setup}
\label{sec:experimental_setup}

\subsection{Training Pipeline}

Training proceeds in two stages. Stage~1 freezes the vision-language backbone and trains the Level-0 decoder and high-resolution RGB branch for 80K steps. Stage~2 initializes from this checkpoint and jointly trains Level-1/Level-2 Depth-VQA and general VQA for 18K steps, with about 60K dense replay examples preserving Level-0 prediction. The dense-to-language interface is active for both instance levels. Because dense predictions and decoder features are detached before geometry-token construction, QA losses update the LoRA-adapted language modules and interface, while dense replay updates the decoder. We use Qwen3.5-4B~\citep{qwen2026qwen35}; Appendix~\ref{app:training_details} provides data composition, preprocessing, optimization, and runtime settings.

\subsection{Evaluation Setup}

For Level~0, we evaluate metric depth on four outdoor datasets (Argoverse~2~\citep{wilson2021argoverse2}, Waymo~\citep{sun2020waymo}, nuScenes~\citep{caesar2020nuscenes}, and DDAD~\citep{guizilini2020ddad}) and five indoor datasets (Taskonomy~\citep{zamir2018taskonomy}, ScanNet++~\citep{yeshwanth2023scannetpp}, HyperSim~\citep{roberts2021hypersim}, SUN RGB-D~\citep{song2015sunrgbd}, and Booster~\citep{ramirez2022booster}). Predictions use ground-truth intrinsics and are scored in metric scale without post-hoc alignment; $\delta_1$ is the primary metric, and the overall score averages the nine dataset-level results.

Depth-VQA follows the tasks and metrics defined in Section~\ref{sec:instance_depth_benchmark}. Level-0 baselines include specialized metric-depth models and depth-aware VLMs; Depth-VQA baselines include general-purpose and depth-aware VLMs. All models use the same prompts and answer protocol. A matched DepthEvidence variant disables the dense-to-language interface at both instance levels to isolate geometry-token conditioning. Appendix~\ref{app:evaluation_protocols} provides the complete baselines, sampling, and scoring protocols.

%% file: sections/06_main_results.tex
\section{Main Results}
\label{sec:main_results}

\subsection{Level-0 Dense Metric Depth Prediction}

\begin{table}[t]
    \centering
    \caption{Level-0 dense metric depth comparison across outdoor and indoor benchmarks. We report $\delta_1\uparrow$ and its unweighted average over all nine datasets. The best and second-best results in each column are shown in bold and underlined, respectively. Rankings are determined using the unrounded values. $^\dagger$DepthLM-12B follows its original sparse point-query protocol~\citep{cai2026depthlm}, using 8,192 randomly sampled pixels per dataset.}
    \label{tab:dense_main}
    \setlength{\tabcolsep}{3.5pt}
    \renewcommand{\arraystretch}{1.08}
    \resizebox{\linewidth}{!}{%
    \begin{tabular}{lcccccccccc}
        \toprule
        \multirow{2}{*}{Method} & \multicolumn{4}{c}{Outdoor} & \multicolumn{5}{c}{Indoor} & \multirow{2}{*}{Avg.} \\
        \cmidrule(lr){2-5}\cmidrule(lr){6-10}
        & Argoverse2 & Waymo & NuScenes & DDAD & Taskonomy & ScanNet++ & HyperSim & SUN RGB-D & Booster & \\
        \midrule
        \rowcolor{gray!15}\multicolumn{11}{c}{\textit{VLMs Trained on Metric Depth Estimation}} \\
        Youtu-VL-4B-Instruct & 0.6290 & 0.5121 & 0.6443 & 0.2995 & 0.7561 & 0.6006 & 0.4285 & 0.3525 & 0.3471 & 0.5077 \\
        DepthVLM-4B & 0.7696 & 0.8056 & 0.8433 & 0.8232 & 0.6992 & 0.8568 & 0.4536 & 0.5089 & 0.2452 & 0.6673 \\
        DepthLM-12B$^\dagger$ & 0.7768 & 0.7758 & 0.8310 & 0.6638 & 0.8218 & 0.7962 & 0.4358 & 0.5508 & 0.5934 & 0.6939 \\
        \rowcolor{gray!15}\multicolumn{11}{c}{\textit{Specialized Vision Models}} \\
        Depth Anything V2 outdoor & 0.1304 & 0.7122 & 0.1701 & 0.1446 & 0.0008 & 0.0002 & 0.0415 & 0.0000 & 0.0000 & 0.1333 \\
        Depth Anything V1 outdoor & 0.2182 & 0.6564 & 0.3338 & 0.1779 & 0.0026 & 0.0001 & 0.0743 & 0.0001 & 0.0000 & 0.1626 \\
        Depth Anything V1 indoor & 0.0018 & 0.0019 & 0.0009 & 0.0015 & 0.4706 & 0.7604 & 0.2504 & 0.6427 & 0.5225 & 0.2947 \\
        Depth Anything V2 indoor & 0.1343 & 0.0188 & 0.1773 & 0.2891 & 0.2397 & 0.2322 & \textbf{0.8652} & 0.1651 & 0.6007 & 0.3025 \\
        Metric3D & 0.7801 & 0.8287 & 0.7691 & 0.8136 & \textbf{0.9467} & 0.0734 & 0.4745 & 0.0582 & 0.0434 & 0.5320 \\
        Metric3D v2 & \underline{0.8589} & \textbf{0.8774} & \underline{0.8555} & 0.8700 & \underline{0.9389} & 0.8809 & 0.7311 & \underline{0.6980} & 0.1374 & 0.7609 \\
        MetricAnything & 0.5920 & \underline{0.8758} & 0.8067 & \textbf{0.8736} & 0.8794 & \textbf{0.8966} & 0.8440 & 0.5713 & 0.5943 & 0.7704 \\
        Depth Anything V3 metric-large & 0.8051 & 0.8483 & 0.7996 & 0.7825 & 0.9189 & \underline{0.8906} & 0.5800 & 0.6141 & \textbf{0.7175} & 0.7730 \\
        UniDepth V2 & \textbf{0.8619} & 0.8502 & \textbf{0.8811} & \underline{0.8719} & 0.8629 & 0.8602 & \underline{0.8648} & 0.3969 & \underline{0.7004} & \underline{0.7945} \\
        \midrule
        \textbf{DepthEvidence-4B (Ours)} & 0.7948 & 0.8266 & 0.8122 & 0.7543 & 0.9072 & 0.8197 & 0.8213 & \textbf{0.8420} & 0.5970 & \textbf{0.7972} \\
        \bottomrule
    \end{tabular}}
\end{table}

\begin{figure}[t]
    \centering
    \includegraphics[width=0.9 \linewidth]{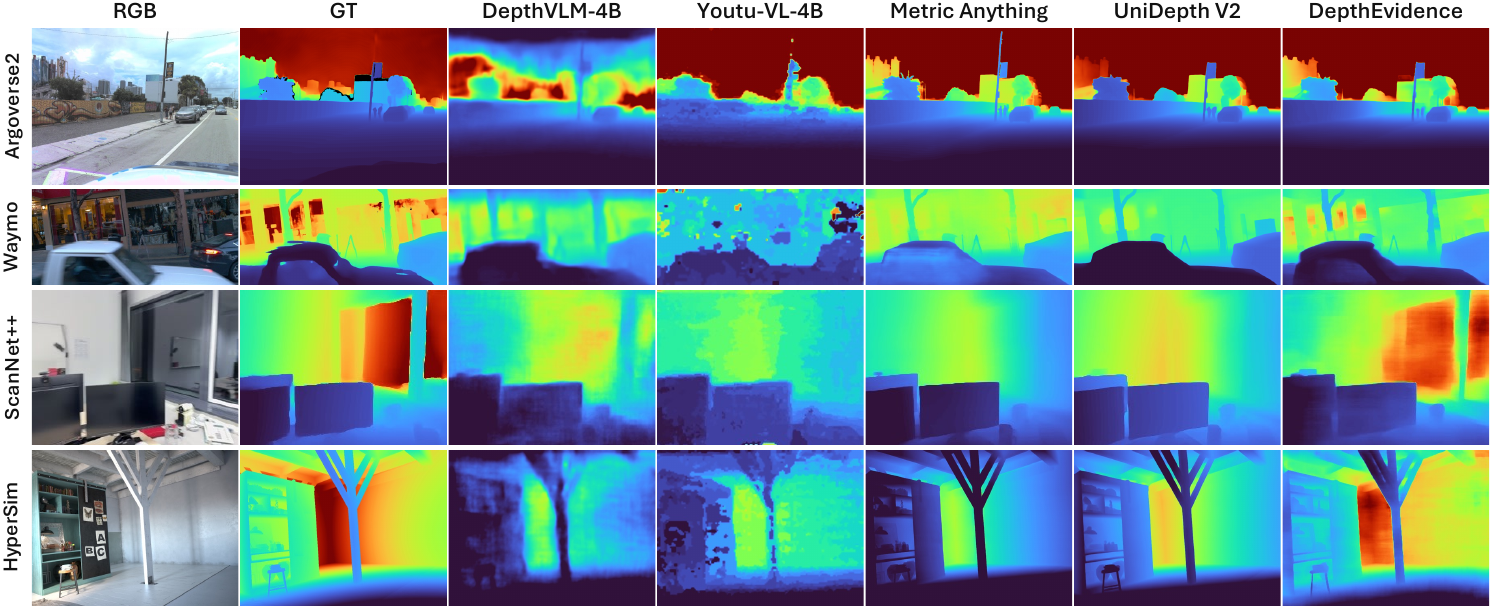}
    \caption{Qualitative dense depth on Argoverse2, Waymo, ScanNet++, and HyperSim. Columns show RGB, ground truth (GT), DepthVLM-4B, Youtu-VL-4B-Instruct, MetricAnything, UniDepth V2, and DepthEvidence. Outdoor GT is densified only for visualization; Table~\ref{tab:dense_main} uses the original sparse sensor GT. Each row shares one color scale; Figure~\ref{fig:dense_metric_qualitative_appendix} shows the other five datasets.}
    \label{fig:dense_metric_qualitative}
\end{figure}

Because general-purpose VLMs lack dense metric-depth outputs, Table~\ref{tab:dense_main} compares DepthEvidence with depth-aware VLMs and specialized estimators. DepthEvidence achieves the highest nine-dataset average $\delta_1$ of 0.7972, slightly above UniDepth V2 (0.7945) and 0.1033 above the strongest depth-aware baseline, DepthLM-12B. It exceeds DepthLM on eight datasets and leads on SUN RGB-D (0.8420). Its outdoor and indoor averages of 0.7970 and 0.7974 indicate balanced aggregate performance, although specialized models remain stronger on several individual datasets.
Figure~\ref{fig:dense_metric_qualitative} complements Table~\ref{tab:dense_main} with two outdoor and two indoor examples; additional visualizations for the remaining five datasets are provided in Appendix Figure~\ref{fig:dense_metric_qualitative_appendix}. DepthEvidence preserves fine structures and depth discontinuities more consistently than DepthVLM-4B and Youtu-VL-4B-Instruct while retaining the global layout. MetricAnything and UniDepth V2 sometimes produce smoother planes, whereas DepthEvidence remains competitive in boundary localization across both domains.

\label{sec:dense_qualitative}

\subsection{Instance-Level Depth Estimation and Reasoning}

\begin{table}[h]
    \centering
    \caption{Instance-level depth estimation and reasoning results. The interface is enabled or disabled at both instance levels. Relative Overall Acc is computed over 370 questions, and Metric Overall Acc includes the 234 questions scored by accuracy. Bold and underlined entries denote the best and second-best results in each column.}
    \label{tab:instance_depth_vqa}
    \setlength{\tabcolsep}{2.5pt}
    \renewcommand{\arraystretch}{1.04}
    \resizebox{\linewidth}{!}{%
    \begin{tabular}{l*{13}{c}}
        \toprule
        \multirow{3}{*}{Method} & \multicolumn{3}{c}{\multirow{2}{*}{Level-1: Instance Metric Depth}} & \multicolumn{10}{c}{Level-2: Compositional Depth Reasoning} \\
        \cmidrule(l){5-14}
        & \multicolumn{3}{c}{} & \multicolumn{4}{c}{Relative} & \multicolumn{6}{c}{Metric} \\
        \cmidrule(lr){2-4}\cmidrule(lr){5-8}\cmidrule(l){9-14}
        & $\delta_1\uparrow$ & RMSE$\downarrow$ & AbsRel$\downarrow$ & \shortstack{Spatial\\Pair} & \shortstack{Spatial\\Group} & Ordering & \shortstack{Overall\\Acc} & \shortstack{Metric Pair\\Acc} & \shortstack{Metric Pair\\$\delta_1\uparrow$} & \shortstack{Spatial Pair\\$\delta_1\uparrow$} & \shortstack{Spatial\\Group} & \shortstack{Set\\Decision} & \shortstack{Overall\\Acc} \\
        \midrule
        \rowcolor{gray!15}\multicolumn{14}{c}{\textit{General-Purpose VLMs}} \\
        Qwen3.5-4B & 0.0159 & 206.1221 & 9.5185 & 0.0930 & 0.0568 & 0.0102 & 0.0405 & 0.6304 & 0.2500 & 0.2234 & 0.3085 & 0.5532 & 0.4701 \\
        Gemini-3.1-Pro-Preview & 0.4108 & 52.4680 & 2.0927 & 0.6395 & 0.5795 & 0.3724 & 0.4838 & 0.6304 & 0.2292 & 0.2340 & 0.3085 & 0.5957 & 0.4872 \\
        Qwen3.7-plus & 0.0766 & 107.2556 & 3.0477 & 0.5814 & 0.6136 & 0.3827 & 0.4838 & 0.6522 & 0.2917 & 0.1489 & 0.3511 & 0.4681 & 0.4573 \\
        Seed-2.1-Turbo & 0.2048 & 107.9360 & 2.7470 & 0.6977 & \underline{0.6250} & 0.4133 & 0.5297 & 0.6304 & 0.1042 & 0.1596 & 0.3617 & 0.4362 & 0.4444 \\
        \rowcolor{gray!15}\multicolumn{14}{c}{\textit{VLMs Trained on Metric Depth Estimation}} \\
        DepthLM-12B & 0.7781 & \underline{10.7460} & \underline{0.1839} & 0.0000 & 0.0000 & 0.0000 & 0.0000 & 0.0000 & 0.1458 & 0.0532 & 0.0000 & 0.0000 & 0.0000 \\
        Youtu-VL-4B-Instruct & 0.0264 & 22.1562 & 1.0852 & 0.1744 & 0.2955 & 0.1224 & 0.1757 & 0.2826 & 0.0417 & 0.1064 & 0.1915 & 0.3404 & 0.2692 \\
        DepthVLM-4B & 0.6592 & 11.9164 & 0.2613 & 0.5000 & 0.3295 & 0.1837 & 0.2919 & 0.5870 & 0.0833 & 0.0957 & 0.3404 & 0.5213 & 0.4615 \\
        \rowcolor{gray!15}\multicolumn{14}{c}{\textit{Agent Pipeline}} \\
        MetricAnything + Qwen3.5-4B & 0.7318 & 11.9224 & 0.2363 & 0.3372 & 0.6136 & 0.4184 & 0.4459 & 0.5870 & 0.2917 & 0.2340 & 0.4149 & 0.6489 & 0.5427 \\
        MetricAnything + Qwen3.7-plus & 0.7543 & 14.2529 & 0.2407 & 0.3140 & \textbf{0.7273} & 0.4541 & 0.4865 & \underline{0.8478} & 0.4167 & 0.2979 & 0.4468 & 0.6596 & 0.6111 \\
        \midrule
        DepthEvidence w/o interface & \underline{0.8653} & 11.2159 & 0.2351 & \underline{0.7442} & 0.5227 & \underline{0.5408} & \underline{0.5838} & 0.8261 & \underline{0.4375} & \underline{0.3511} & \underline{0.7447} & \underline{0.7660} & \underline{0.7692} \\
        \textbf{DepthEvidence (Ours)} & \textbf{0.8666} & \textbf{8.6062} & \textbf{0.1629} & \textbf{0.7907} & 0.5568 & \textbf{0.5867} & \textbf{0.6270} & \textbf{0.9130} & \textbf{0.5833} & \textbf{0.4043} & \textbf{0.8191} & \textbf{0.8511} & \textbf{0.8504} \\
        \bottomrule
    \end{tabular}}
\end{table}

Table~\ref{tab:instance_depth_vqa} evaluates instance-level metric estimation and compositional reasoning. Among the depth-aware baselines, DepthLM-12B~\citep{cai2026depthlm} performs best on Level~1 ($\delta_1=0.7781$), whereas DepthVLM-4B~\citep{yu2026depthvlm} performs best on both Level~2 tracks (0.2919 Relative and 0.4615 Metric Overall Acc). DepthEvidence leads all evaluated methods on Level~1, with $\delta_1=0.8666$, RMSE $=8.6062$, and AbsRel $=0.1629$. It also achieves 0.6270 Relative and 0.8504 Metric Overall Acc, exceeding Seed-2.1-Turbo and Gemini-3.1-Pro-Preview by 9.73 and 36.32 percentage points, respectively, and DepthVLM-4B by 33.51 and 38.89 points. DepthEvidence further obtains the highest reported numerical-gap $\delta_1$ on Metric Pair (0.5833) and Spatial Metric Pair (0.4043).

\subsection{Effect of the Dense-to-Language Interface}
\label{sec:dense_to_language_results}

\begin{figure}[t]
    \centering
    \includegraphics[width=0.9 \linewidth]{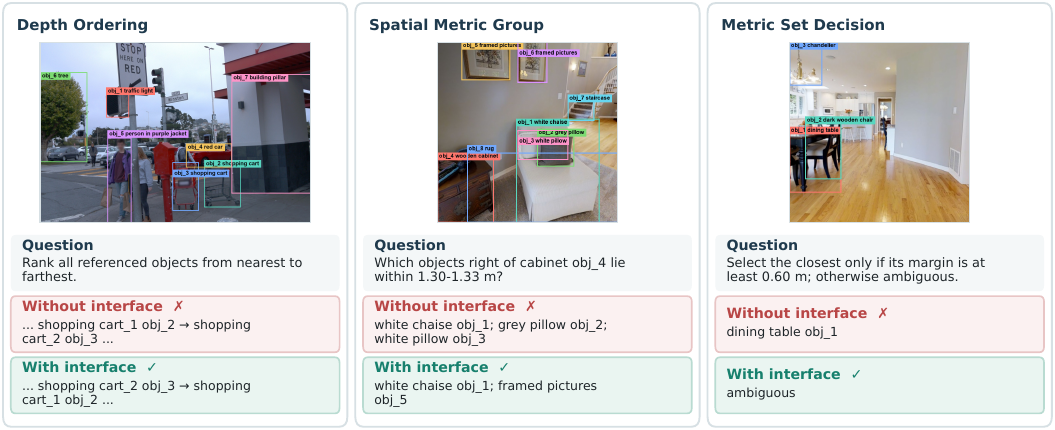}
    \caption{Representative Level-2 cases corrected by the dense-to-language interface. The three panels test complete depth ordering, spatially filtered metric-range selection, and margin-aware ambiguity decisions, respectively. Each panel contrasts matched variants without and with the interface; green answers match the ground truth. Aggregate results are reported in Table~\ref{tab:instance_depth_vqa}.}
    \label{fig:interface_cases}
\end{figure}

Matched ablations in Table~\ref{tab:instance_depth_vqa} isolate the interface effect. On Level~1, $\delta_1$ is nearly unchanged (0.8653$\rightarrow$0.8666), while RMSE and AbsRel decrease from 11.2159 to 8.6062 and from 0.2351 to 0.1629, suggesting fewer large metric errors. On Level~2, Overall Acc increases by 4.32 points on Relative and 8.12 points on Metric. The largest categorical gain is on Metric Pair (+8.69 points), while numerical-gap $\delta_1$ rises from 0.4375 to 0.5833 on Metric Pair and from 0.3511 to 0.4043 on Spatial Metric Pair. Although MetricAnything + Qwen3.7-plus leads Relative Spatial Group (0.7273), its Relative and Metric Overall scores (0.4865/0.6111) trail DepthEvidence by 14.05/23.93 points, and both agent pipelines remain below DepthEvidence without the interface (Appendix~\ref{app:agent_pipeline}). Overall, the agent pipelines lag the full DepthEvidence system, indicating that external estimator--VLM coupling is less effective than our integrated design.
Figure~\ref{fig:interface_cases} visualizes representative corrections in depth ordering, spatially filtered metric-range selection, and margin-aware ambiguity decisions.

\subsection{General VQA Retention and Spatial Understanding}
\label{sec:general_vqa_results}

\begin{table}[t]
    \centering
    \caption{General and spatial capability evaluation. Metrics follow their native scales; all are higher-is-better except Ego3D RMSE. Bold marks the best result, and ``--'' denotes unavailable results.}
    \label{tab:general_vqa}
    \scriptsize
    \setlength{\tabcolsep}{3.8pt}
    \renewcommand{\arraystretch}{0.82}
    \setlength{\aboverulesep}{0.6pt}
    \setlength{\belowrulesep}{0.8pt}
    \resizebox{0.97\linewidth}{!}{%
    \begin{tabular}{lcccccccc}
        \toprule
        \multirow{2}{*}{Method} & \multicolumn{4}{c}{STEM} & \multicolumn{4}{c}{General VQA} \\
        \cmidrule(lr){2-5}\cmidrule(l){6-9}
        & MathVision & MathVista & MMMU & MMMU-Pro & RealWorldQA & MMStar & MMBench & HallusionBench \\
        \midrule
        Youtu-VL-4B-Instruct & -- & 76.50 & 61.10 & 43.00 & 74.60 & 71.10 & 83.90 & 59.10 \\
        GLM-4.1V-9B-Thinking & 54.40 & 80.70 & 68.00 & 57.10 & -- & 72.90 & 85.80 & 63.20 \\
        \midrule
        Qwen3.5-4B & \textbf{74.60} & \textbf{85.10} & 77.60 & \textbf{66.30} & \textbf{79.50} & \textbf{78.30} & 89.40 & 65.00 \\
        \textbf{DepthEvidence (Ours)} & 74.10 & 84.90 & \textbf{78.71} & 64.05 & 78.06 & \textbf{78.30} & \textbf{89.52} & \textbf{68.21} \\
        \bottomrule
    \end{tabular}}
    \par\vspace{2pt}\nointerlineskip
    \resizebox{0.97\linewidth}{!}{%
    \begin{tabular}{lcccccccc}
        \toprule
        \multirow{2}{*}{Method} & \multicolumn{4}{c}{Text Recognition and Document Understanding} & \multicolumn{4}{c}{Spatial Intelligence} \\
        \cmidrule(lr){2-5}\cmidrule(l){6-9}
        & AI2D & CC-OCR & OCRBench & CharXiv (RQ) & ERQA & EmbSpatialBench & Ego3D Acc. & Ego3D RMSE$\downarrow$ \\
        \midrule
        Youtu-VL-4B-Instruct & 85.60 & -- & 81.30 & 43.80 & -- & -- & 45.63 & 19.27 \\
        GLM-4.1V-9B-Thinking & 87.90 & -- & 84.20 & -- & 45.80 & -- & 56.93 & 12.79 \\
        \midrule
        Qwen3.5-4B & 89.60 & 76.70 & \textbf{85.00} & \textbf{70.80} & 54.00 & 81.30 & 62.82 & 13.17 \\
        \textbf{DepthEvidence (Ours)} & \textbf{90.67} & \textbf{77.34} & 84.60 & 69.00 & \textbf{57.25} & \textbf{83.14} & \textbf{67.99} & \textbf{6.83} \\
        \bottomrule
    \end{tabular}}
\end{table}

Table~\ref{tab:general_vqa} shows that DepthEvidence improves spatial reasoning while largely retaining the general capabilities of its Qwen3.5-4B initialization. Non-spatial scores change modestly: MMMU, MMBench, HallusionBench, AI2D, and CC-OCR improve, while the largest regression is 2.25 points on MMMU-Pro. All four spatial metrics improve: ERQA rises from 54.00 to 57.25, EmbSpatialBench from 81.30 to 83.14, and Ego3D accuracy from 62.82 to 67.99, while RMSE falls from 13.17 to 6.83 ($-48.1\%$). DepthEvidence also outperforms Youtu-VL-4B-Instruct~\citep{youtu2026vl} and GLM-4.1V-9B-Thinking~\citep{glm2025glm41v} on their commonly reported higher-is-better metrics and Ego3D RMSE.

%% file: sections/07_analysis.tex
\section{Analysis and Discussion}
\label{sec:analysis}

\subsection{Dense Decoder Ablation}

\begin{table}[t]
    \centering
    \begin{minipage}[t]{0.47\linewidth}
        \centering
        \caption{Dense-decoder ablation on outdoor and indoor datasets ($\delta_1\uparrow$).}
        \label{tab:dense_decoder_ablation}
        \setlength{\tabcolsep}{4pt}
        \renewcommand{\arraystretch}{0.9}
        \resizebox{0.96\linewidth}{!}{%
        \begin{tabular}{cccccc}
            \toprule
            \multirow{2}{*}{SSI} & \multirow{2}{*}{HR branch} & \multicolumn{2}{c}{Outdoor} & \multicolumn{2}{c}{Indoor} \\
            \cmidrule(lr){3-4}\cmidrule(lr){5-6}
            & & Argoverse2 & Waymo & Taskonomy & ScanNet++ \\
            \midrule
            -- & -- & 0.7313 & 0.7505 & 0.7852 & 0.7349 \\
            $\checkmark$ & -- & 0.7547 & 0.7736 & 0.8175 & 0.7511 \\
            -- & $\checkmark$ & 0.7824 & 0.8129 & 0.8804 & 0.8081 \\
            $\checkmark$ & $\checkmark$ & \textbf{0.7948} & \textbf{0.8266} & \textbf{0.9072} & \textbf{0.8197} \\
            \bottomrule
        \end{tabular}}
    \end{minipage}\hfill
    \begin{minipage}[t]{0.51\linewidth}
        \centering
        \caption{Geometry-interface ablation on Level-2 Relative and Metric Overall Acc.}
        \vspace{2pt}
        \label{tab:interface_ablation}
        \setlength{\tabcolsep}{2.4pt}
        \renewcommand{\arraystretch}{0.9}
        \resizebox{0.98\linewidth}{!}{%
        \begin{tabular}{lcccccc}
            \toprule
            Variant & $\mathcal{L}_{\mathrm{ans}}$ & $\mathcal{L}_{\mathrm{raw}}$ & $\mathcal{L}_{\mathrm{ctx}}$ & $\mathcal{L}_{\mathrm{rank}}$ & Relative Overall & Metric Overall \\
            \midrule
            w/o interface & -- & -- & -- & -- & 0.5838 & 0.7692 \\
            \multirow{4}{*}{Interface} & $\checkmark$ & -- & -- & -- & 0.5838 & 0.7735 \\
            & $\checkmark$ & $\checkmark$ & -- & -- & 0.5946 & 0.8120 \\
            & $\checkmark$ & $\checkmark$ & $\checkmark$ & -- & 0.6135 & 0.8419 \\
            & $\checkmark$ & $\checkmark$ & $\checkmark$ & $\checkmark$ & \textbf{0.6270} & \textbf{0.8504} \\
            \bottomrule
        \end{tabular}}
    \end{minipage}
\end{table}

Table~\ref{tab:dense_decoder_ablation} shows that both the SSI objective and high-resolution branch increase $\delta_1$ on each evaluated dataset, with their combination obtaining the highest values among the ablated variants. The HR branch has a larger measured effect indoors: used alone, it raises mean $\delta_1$ by 0.0842 on the two indoor datasets versus 0.0568 outdoors. One possible explanation is the sparse evaluation coverage of outdoor datasets: their ground truth contains fewer evaluated pixels near object boundaries, so improvements in boundary prediction contribute less to the measured $\delta_1$ than under denser indoor annotations.

\subsection{Geometric Supervision Ablation}

The interface is used at both instance levels; Table~\ref{tab:interface_ablation} focuses on its supervision design using the two Level-2 accuracy aggregates. Training the interface with only $\mathcal{L}_{\mathrm{ans}}$ leaves Relative Overall Acc unchanged at 0.5838 and raises Metric Overall Acc from 0.7692 to 0.7735. Adding $\mathcal{L}_{\mathrm{raw}}$ improves the two aggregates by 1.08 and 3.85 percentage points, respectively; $\mathcal{L}_{\mathrm{ctx}}$ adds a further 1.89 and 2.99 points, and $\mathcal{L}_{\mathrm{rank}}$ adds 1.35 and 0.85 points. The complete objective reaches 0.6270 Relative Overall Acc and 0.8504 Metric Overall Acc, improving over the model without the interface by 4.32 and 8.12 percentage points. These results support the utility of auxiliary geometric supervision, with a larger measured gain on the accuracy-based Metric tasks. A separate no-interface control with direct metric supervision remains clearly below the full interface, indicating that the main performance gain arises from the interface itself rather than numerical supervision alone; Appendix~\ref{app:interface_metric_ablation} provides the detailed disentanglement analysis.

\subsection{Limitations}
\label{sec:limitations}

The current setting assumes access to camera intrinsics and explicit object references. Its geometric evidence is therefore sensitive to errors in calibration, object grounding, and dense-depth prediction, particularly when depth ranges or decision margins are narrow. In addition, although cached continuation avoids repeated prefix computation, conditioning on geometry tokens incurs additional latency and memory relative to inference without the interface (Appendix~\ref{app:inference_efficiency}).

%% file: sections/08_conclusion.tex
\section{Conclusion}
\label{sec:conclusion}

DepthEvidence connects dense metric depth prediction, instance-level metric depth estimation, and compositional depth reasoning within a shared MLLM. Its dense-to-language interface supplies object-aligned continuous geometry tokens, with geometric supervision encouraging metric-depth recoverability during language-context interaction. Experiments show competitive dense prediction and gains in object measurement, numerical depth gaps, and metric-constrained decisions. General VQA remains broadly stable relative to the base model, while spatial understanding improves on the evaluated benchmarks. These findings support dense prediction as both a supervised output and internal evidence for language reasoning.
Future work can assess robustness to noisier camera intrinsics and object references, reduce inference cost as the object count grows, and extend the interface to other geometric evidence.

%% file: sections/09_appendix.tex
\section{Method and Implementation Details}
\label{app:method_details}

\subsection{Dense Depth Branch}

\subsubsection{Decoder and Camera Conditioning}
\label{app:dense_decoder}

Given an RGB image $\mathbf{I}\in\mathbb{R}^{H\times W\times 3}$, we retain three intermediate feature maps $\{\mathbf{V}_k\}_{k=1}^{3}$ from the vision encoder and reshape the final VLM hidden states at image-token positions into a spatial map $\mathbf{M}$. Following DPT~\citep{ranftl2021vision}, we normalize and project these four maps to $\{256,512,1024,1024\}$ channels, resample them with strides $\{8,4,2,1\}$, and map each level to a common width of 512 channels. The camera condition is applied to all four levels before fusion, while RGB features are injected into the three higher-resolution levels. Residual fusion blocks propagate the coarsest representation upward and merge it with progressively finer conditioned features. The resulting full-resolution feature is used both to predict dense metric depth and to construct object-level geometry representations.

Let $\boldsymbol{\kappa}=(f_x,f_y,c_x,c_y)$ denote the supplied camera intrinsics for an image of original size $W_0\times H_0$. We convert them to image-relative coordinates $\bar{\mathbf{k}}=[\bar f_x,\bar f_y,\bar c_x,\bar c_y]=[f_x/W_0,f_y/H_0,c_x/W_0,c_y/H_0]$, and define $\phi_x=2\arctan(1/(2\bar f_x))$, $\phi_y=2\arctan(1/(2\bar f_y))$, and $r=W_0/H_0$. The decoder condition is the nine-dimensional vector
\begin{equation}\mathbf{s}(\boldsymbol{\kappa})=[\bar{\mathbf{k}},\log\bar f_x,\log\bar f_y,\phi_x,\phi_y,r].\label{eq:camera_condition}\end{equation}
An MLP encodes $\mathbf{s}(\boldsymbol{\kappa})$ as $\mathbf{c}$ and modulates each decoder level through $\operatorname{FiLM}(\mathbf{F}_k,\mathbf{c})=\mathbf{F}_k\odot(1+\boldsymbol{\gamma}_k(\mathbf{c}))+\boldsymbol{\beta}_k(\mathbf{c})$.

\subsubsection{High-resolution RGB Refinement}
\label{app:rgb_refinement}

The RGB branch processes the input image with a convolutional stem followed by three stride-2 stages, producing feature maps $\{\mathbf{R}_k\}_{k=1}^{3}$ at strides $\{2,4,8\}$. After channel projection and spatial alignment, each RGB map is added to its corresponding DPT level through a learned residual spatial gate, $\widetilde{\mathbf{F}}_k=\mathbf{F}_k+\mathbf{G}_k\odot\mathbf{R}_k$, where $\mathbf{G}_k=\operatorname{clip}(b_k+1.5\tanh g_k([\mathbf{F}_k;\mathbf{R}_k]),0,3)$. The function $g_k$ predicts a spatially varying residual from the concatenated DPT and RGB features, and the base gate $b_k$ is initialized to $0.01$. After the final fusion block, a convolution reduces the feature width, a second FiLM layer applies the camera condition, and the feature map is bilinearly resized to the target resolution. The metric prediction is $\widehat{\mathbf{D}}=\operatorname{clip}(\operatorname{softplus}(h(\mathbf{F}))+d_{\min},d_{\min},d_{\max})$, with $d_{\min}=0.05$ m and $d_{\max}=250$ m.

\subsubsection{Dense Objective}
\label{app:dense_objective}

The dense objective combines the absolute relative and scale-invariant log losses~\citep{eigen2014depth} with structural supervision on inverse depth. Specifically, $\mathcal{L}_{\mathrm{der}}$ and $\mathcal{L}_{\mathrm{lap}}$ match Scharr derivatives and Laplacian responses at scales $\{1,2,4,8\}$. For the edge-guided stable local scale-and-shift invariant term $\mathcal{L}_{\mathrm{ssi}}$, we sample regions centered on strong RGB edges and transform depth using $q(D)=\log(1+50/(10^{-4}+D))$. Prediction and target values are robustly normalized and compared both within the sampled local regions and over the full image.

\subsection{Depth-VQA Instructions and Grounding}
\label{app:instance_instruction}

A Level-1 example contains an image $\mathbf{I}$ and a grounded target object $o_i$, and asks for instance-level metric depth estimation. The default target $d_i$ is the mean of valid depths within the object's instance mask; queries can also request center-point, median, nearest, or farthest depth. A Level-2 example contains grounded objects $\mathcal{O}=\{o_i\}_{i=1}^{N}$ and a question $q$ requiring compositional depth reasoning in the Relative or Metric track. Relative questions ask for pairwise comparisons, closest or farthest object selection, or depth ordering. Metric questions ask for numerical mean-depth gaps, comparisons against a threshold, selection within a depth range or bounds, or decisions under a required depth margin. Compositional questions combine these operations with image-space relations to select the relevant objects. For a margin-based closest-object decision, the target is \emph{ambiguous} when the closest object fails to meet the specified margin.

\paragraph{Depth convention and object statistics.}
Camera depth denotes the optical-axis coordinate $Z$ of a point in the camera frame, measured in meters, rather than its Euclidean distance to the camera center. Object depth denotes the statistic requested for an instance. For Level~1, let $\Omega_i$ contain the pixels inside the ground-truth instance mask with valid camera depth $D(p)$. The mean target is $d_i^{\mathrm{mean}}=|\Omega_i|^{-1}\sum_{p\in\Omega_i}D(p)$, and the median target is the median over the same set. The nearest and farthest targets are the 5th and 95th percentiles of valid mask depths, respectively; center-point depth is a point measurement rather than a region aggregate. The mask defines the target statistics, while a box or referring prompt identifies the object to the model. The interface pools predicted evidence over box regions (Appendix~\ref{app:geometry_interface_details}). Mean camera-depth gaps compare optical-axis depth statistics, not Euclidean distances between objects in 3D.

A box prompt specifies the object category and bounding box, with coordinates normalized to a $0$--$1000$ image coordinate system. A referring prompt identifies the object with a natural-language description enclosed by object-reference tokens. Each object in a Level-2 prompt receives a stable identifier such as \texttt{obj\_1}, and the prompt lists its identifier, category, and grounding cue before the question.

Every Depth-VQA instruction embeds the camera intrinsics in the user prompt as \texttt{<focal> $f_x$ $f_y$ $c_x$ $c_y$ </focal>}. The token name \texttt{<focal>} is retained in the implementation, but the block contains both focal lengths and principal-point coordinates. The language-modeling loss is applied only to the final answer: a metric depth value for Level~1, or a numerical depth gap, an object or set selection, an ordering, or a discrete comparison or constraint decision for Level~2, including \emph{ambiguous} for an unmet depth-margin requirement. Input positions are excluded as language-modeling targets; auxiliary geometric losses supervise the geometry representations as specified in Appendix~\ref{app:geometry_objectives}.

\subsection{Dense-to-Language Interface}

\subsubsection{Construction}
\label{app:geometry_interface_details}

The same dense-to-language interface is used for Level~1 single-object queries and Level~2 multi-object questions. The dense decoder produces a metric depth map $\widehat{\mathbf{D}}$ and a full-resolution fused feature map $\mathbf{F}^{d}$. For each grounded object $o_i$, let $\mathbf{b}_i=[x_i^1,y_i^1,x_i^2,y_i^2]$ be its normalized box and let $\mathcal{R}_i$ denote the corresponding region. We average-pool $\mathbf{F}^{d}$ inside $\mathcal{R}_i$ to obtain an appearance-and-shape descriptor $\mathbf{r}_i$, and summarize the predicted depths in the same region as $\mathbf{s}_i=[\mu_i,\sigma_i,d_i^{\min},d_i^{\max},d_i^{\mathrm{ctr}},\log\mu_i,a_i,\rho_i]$. These entries denote the mean, standard deviation, minimum, maximum, center depth, log-mean depth, box area, and aspect ratio, respectively.

For a box-grounded object, we directly use the box given in the prompt. For a referring expression enclosed by \texttt{<|object\_ref\_start|>} and \texttt{<|object\_ref\_end|>}, we take the VLM hidden state at the closing reference token as an object query. A lightweight box-regression head, implemented as layer normalization followed by a two-layer MLP, predicts $\widehat{\mathbf{b}}_i\in[0,1]^4$. The predicted box is supervised by the ground-truth region during training and is used for differentiable region pooling and inference.

The projected object tokens are organized as
\begin{equation}\mathcal{B}_{G}=[\mathbf{g}_{\mathrm{start}},e(\texttt{obj\_1:}),\mathbf{g}_{1,1},\mathbf{g}_{1,2},\ldots,e(\texttt{obj\_N:}),\mathbf{g}_{N,1},\mathbf{g}_{N,2},\mathbf{g}_{\mathrm{end}}],\label{eq:geometry_bank}\end{equation}
where $e(\texttt{obj\_i:})$ is the ordinary language-embedding sequence of an object anchor and $\mathbf{g}_{\mathrm{start}}$ and $\mathbf{g}_{\mathrm{end}}$ are learned boundary vectors. Placeholder \texttt{<geometry>} embeddings at the dynamic positions are replaced by the corresponding object vectors before the continuation enters the language model. The bank therefore contains $2N+2$ dynamically generated vectors for $N$ objects, in addition to the ordinary textual anchors. Its positions serve as conditioning context and are excluded as language-modeling targets. The answer loss nevertheless backpropagates through the geometry representations to the projector. The bank contains predicted geometry; ground-truth depth statistics serve as targets for the auxiliary representation objectives.

\subsubsection{Token Capacity: Why Two per Object?}
\label{app:two_geometry_tokens}

We use $K=2$ as a compact multi-token bottleneck for each grounded object. The fused descriptor $[\mathbf{r}_i;\mathbf{b}_i;\mathbf{s}_i]$ contains decoder features, image location, and several statistics of the predicted metric depth. Compressing all of this evidence into a single VLM vector would impose a narrower information bottleneck. Instead, the final layer of $P_g$ produces a $2h$-dimensional output that is reshaped into two independently parameterized vectors in $\mathbb{R}^{h}$. Both vectors are computed from the complete object descriptor; we do not prescribe one token for appearance and the other for metric depth. A shared object-index embedding binds the pair to the same instance, while distinct slot embeddings break their symmetry and allow the language model to use the two positions differently. The resulting specialization, if any, is therefore learned rather than hand-crafted. This added representational bandwidth requires only one additional dynamic position per object relative to $K=1$: for $N$ grounded objects, the geometry bank contains $2N$ object-geometry positions instead of $N$. Thus, $K=2$ provides a controlled capacity increase while keeping the interface sequence length proportional to the number of grounded objects; the subsequent attention cost follows the standard dependence on the resulting total sequence length.

\subsubsection{Cached Geometry Continuation}
\label{app:cached_geometry_details}

At both instance levels, let $\mathbf{P}$ contain the image, camera-intrinsics block, grounded object descriptions, and question. Processing this prefix once produces contextualized states $\mathbf{H}_{P}$ and a layer-wise cache $\mathcal{C}_{P}$. The dense branch uses the visual positions in $\mathbf{H}_{P}$ together with the multi-level vision features to construct $\mathcal{B}_{G}$, after which the cache is extended and the answer is generated as
\begin{equation}(\mathbf{H}_{P},\mathcal{C}_{P})=f_{\theta}(\mathbf{P};\operatorname{cache}=\mathrm{on}),\qquad \mathcal{C}_{P,G}=f_{\theta}(\mathcal{B}_{G};\mathcal{C}_{P}),\qquad \mathbf{Y}\sim f_{\theta}(\cdot;\mathcal{C}_{P,G}).\label{eq:cached_geometry_continuation}\end{equation}
In a full-attention layer, $\mathcal{C}_{P}$ contains the prefix keys and values; queries from the geometry continuation attend to these cached tensors and to preceding continuation positions. The language backbone also contains linear-attention layers, which summarize the prefix with a causal-convolution state and a recurrent state rather than a conventional key--value sequence. We clone these differentiable prefix states and use them to initialize the chunk kernels that process the multi-token geometry continuation. Both attention types therefore process the continuation as if $[\mathbf{P};\mathcal{B}_{G};\mathbf{Y}]$ had been evaluated as one causal sequence. Caching changes the computation schedule without changing causal order or exposing the geometry tokens as visible outputs.

\subsubsection{Representation Objectives}
\label{app:geometry_objectives}

For object $i$, let $\bar{\mathbf{g}}_i=(\mathbf{g}_{i,1}+\mathbf{g}_{i,2})/2$ be the raw representation before it enters the language model, and let $\bar{\mathbf{h}}_i=(\mathbf{h}_{i,1}+\mathbf{h}_{i,2})/2$ be the final-layer representation at the same geometry positions after cached continuation. Let $\mathbf{d}_i$ contain five ground-truth object-depth statistics: center, mean, median, nearest, and farthest depth. Two lightweight heads predict their logarithms from the raw and contextualized representations:
\begin{equation}\widehat{\mathbf{z}}^{\mathrm{raw}}_i=H_{\mathrm{raw}}(\bar{\mathbf{g}}_i),\qquad \widehat{\mathbf{z}}^{\mathrm{ctx}}_i=H_{\mathrm{ctx}}(\bar{\mathbf{h}}_i),\qquad \mathbf{z}_i=\log\mathbf{d}_i.\label{eq:geometry_depth_heads}\end{equation}
The valid-depth ratio $w_i$ weights the corresponding smooth-$\ell_1$ regressions,
\begin{equation}\mathcal{L}_{\mathrm{raw}}=\frac{\sum_i w_i\,\operatorname{SmoothL1}(\widehat{\mathbf{z}}^{\mathrm{raw}}_i,\mathbf{z}_i)}{\sum_i w_i},\qquad \mathcal{L}_{\mathrm{ctx}}=\frac{\sum_i w_i\,\operatorname{SmoothL1}(\widehat{\mathbf{z}}^{\mathrm{ctx}}_i,\mathbf{z}_i)}{\sum_i w_i}.\label{eq:geometry_depth_losses}\end{equation}
At both instance levels, we operationalize metric information preservation through the recoverability of these ground-truth statistics from both representations. The raw and contextual losses supervise the projected tokens and their states after language-context interaction against the same targets, encouraging metric content to remain accessible to a prediction head. For multi-object examples, we additionally supervise relative depth using
\begin{equation}\mathcal{L}_{\mathrm{rank}}=\frac{1}{|\mathcal{P}|}\sum_{(i,j)\in\mathcal{P}}\operatorname{softplus}\!\left(-s_{ij}(\widehat{z}_i-\widehat{z}_j)\right),\qquad s_{ij}=\operatorname{sign}(\log d_i^{\mathrm{mean}}-\log d_j^{\mathrm{mean}}),\label{eq:geometry_rank_loss}\end{equation}
where $\widehat{z}_i$ is the contextual prediction of mean log-depth and $\mathcal{P}$ contains object pairs whose ground-truth log-depth gap exceeds a small ambiguity margin. The differentiable cache allows the answer and contextual geometry objectives to update the continuation adapters and trainable prefix-pathway parameters. The geometry projector receives gradients from the answer, raw-depth, contextual-depth, and ranking losses. We detach $\widehat{\mathbf{D}}$ and $\mathbf{F}^{d}$ before object projection, so these objectives do not perturb the pretrained dense decoder, which continues to be optimized by $\mathcal{L}_{\mathrm{dense}}$ on dense-prediction batches.

\section{Data and Training Details}
\label{app:training_details}

\subsection{Data Composition}
The Level-0 corpus combines approximately 6.7M images from LingBot-Depth, Taskonomy, Waymo, DDAD, nuScenes, Argoverse~2, and ScanNet++. We apply depth completion to outdoor sources with sparse sensor depth so that every source image is paired with a dense depth map, and preserve the camera intrinsics associated with each image. The dense branch converts these intrinsics into its numerical camera condition, whereas Depth-VQA prompts provide the raw $(f_x,f_y,c_x,c_y)$ values in a textual camera-intrinsics block. All dense-benchmark frames are excluded from training. The instance-level corpus contains approximately 1.5M Level-1 and Level-2 examples sampled from LingBot-Depth, Waymo, DDAD, Argoverse~2, nuScenes, and Taskonomy. Stage~2 combines these examples with roughly 771K general visual-instruction examples and about 60K Level-0 replay samples, with stream frequencies proportional to corpus size.

\subsection{Optimization and Runtime}

\paragraph{Optimization.} We use Qwen3.5-4B as the vision-language backbone and AdamW with weight decay 0.01 and a 3\% warmup ratio. During Stage~1, the vision encoder and language model are frozen, and the dense branch is trained with a learning rate of $10^{-3}$. During Stage~2, decoder-side modules and LoRA-adapted language parameters use a learning rate of $10^{-4}$, while the vision encoder uses $10^{-7}$. LoRA is applied to attention and feed-forward projections with rank 32, scaling factor 64, and dropout 0.05.

\paragraph{Runtime settings.} Dense inputs are resized to $448\times448$. Instruction-tuning sequences contain at most 4,096 tokens, including up to 2,048 visual tokens. We train in bfloat16 with FlashAttention-2 and an effective global batch size of 256 for both dense and VQA updates. Stage~2 runs on four nodes with 32 NVIDIA A100 GPUs in total.

\section{Benchmark Construction and Evaluation Protocols}
\label{app:evaluation_protocols}

\subsection{Benchmark Construction and Quality Control}
\label{app:benchmark_construction}

\paragraph{Dense depth preparation.}
Outdoor datasets provide sparse sensor measurements, so we first apply depth completion to obtain a dense depth map for every source image. These image--depth pairs form the Level-0 pool and provide the metric field from which instance-level targets are derived.

\paragraph{Level-1 instance construction.}
For each image, Qwen3.7-plus~\citep{qwen2026qwen37plus} and Seed-2.1-Turbo~\citep{bytedance2026seed21} independently identify the major objects and propose a category and grounding box for each one. We pool their candidates and discard boxes covering less than 1\% of the image or lying too close to image boundaries. The remaining category--box pairs prompt SAM~3~\citep{carion2025sam3} to produce instance masks, which pass a second mask-quality filter. For each retained mask, we combine its label, box, and dense depth values to compute center-point, mean, median, nearest, and farthest targets; the latter two are the 5th and 95th percentiles of valid mask depths, respectively. The resulting grounded instance records constitute the Level-1 data. Using two proposal models reduces dependence on model-specific object and category preferences during automatic construction.

\paragraph{Level-2 question construction.}
We sample multiple Level-1 instance records from an image according to the Relative and Metric task taxonomy in Table~\ref{tab:level2_benchmark_full}. Given the selected object labels, boxes, and depth statistics, either Qwen3.7-plus or Seed-2.1-Turbo generates a candidate question and answer. The other model then checks question validity and answer correctness, and invalid examples are removed. This cross-model generation--verification protocol reduces reliance on the phrasing and reasoning patterns of a single generator.

\paragraph{Human review and distribution control.}
We manually inspect 1\% of the training samples across Levels~0--2, aggregate recurring failure cases, and translate them into targeted filters applied to the training corpus. Every benchmark example undergoes manual verification. For video data, we favor different objects when constructing Level-1 and Level-2 questions within the same scene and limit reuse of the same instance across task types, reducing repeated-instance bias.

\subsection{Benchmark Composition and Partitioning}

\begin{table}[!htbp]
    \centering
    \caption{Detailed composition of the Relative Depth and Metric Depth tracks in the Level-2 benchmark. ``Subset'' orders a specified subset of objects, whereas ``all'' orders every referenced object.}
    \label{tab:level2_benchmark_full}
    \scriptsize
    \setlength{\tabcolsep}{2.5pt}
    \renewcommand{\arraystretch}{0.96}
    \begin{tabular}{@{}p{0.10\linewidth}p{0.17\linewidth}p{0.25\linewidth}p{0.40\linewidth}@{}}
        \toprule
        Track & Subbench & Raw question type & Required reasoning \\
        \midrule
        \multirow{8}{*}[-20pt]{\shortstack[c]{Relative\\Depth}} & \multirow{3}{=}{Spatial Pair} & \path{closer_farther} & Decide which of two objects is closer to or farther from the camera. \\
        & & \path{front_behind} & Decide which of two objects is in front of or behind the other. \\
        & & \path{spatial_depth} & Resolve a left/right reference, then compare the selected object with another object in depth. \\
        \cmidrule(l){2-4}
        & Spatial Group & \path{spatial_depth_group} & Filter objects by a spatial relation, then select the nearest or farthest member. \\
        \cmidrule(l){2-4}
        & \multirow{4}{=}{Depth Ordering} & \path{depth_ordering_closest} & Select the closest object from all candidates. \\
        & & \path{depth_ordering_farthest} & Select the farthest object from all candidates. \\
        & & \path{depth_ordering_subset} & Order a specified subset of objects by depth. \\
        & & \path{depth_ordering_all} & Order all referenced objects by depth. \\
        \midrule
        \multirow{7}{*}[-30pt]{\shortstack[c]{Metric\\Depth}} & \multirow{2}{=}{Metric Pair} & \path{metric_depth_gap} & Compute the numerical difference in mean camera depth between two objects. \\
        & & \path{metric_gap_threshold} & Determine whether the mean-depth gap between two objects reaches a specified metric threshold. \\
        \cmidrule(l){2-4}
        & Spatial Metric Pair & \path{spatial_metric_gap} & Resolve a 2D spatial reference, then compute the absolute mean-depth difference to another object. \\
        \cmidrule(l){2-4}
        & \multirow{2}{=}{Spatial Metric Group} & \path{spatial_range_filter} & Apply a spatial filter, then select objects whose mean depths fall inside a specified metric range. \\
        & & \path{compositional_metric_constraint} & Combine a spatial condition, a metric lower bound, and an object-referenced upper-depth bound. \\
        \cmidrule(l){2-4}
        & \multirow{2}{=}{Metric Set Decision} & \path{camera_safety_band} & Determine whether any referenced object lies within a specified mean-depth range from the camera. \\
        & & \path{metric_margin_decision} & Select the closest object only when its depth advantage exceeds a metric margin; otherwise answer \emph{ambiguous}. \\
        \bottomrule
    \end{tabular}
\end{table}

\needspace{0.30\textheight}
\paragraph{Benchmark partitioning.} For datasets with official partitions, benchmark examples are sampled exclusively from their official validation or test splits. For datasets without an official evaluation partition, we randomly hold out complete scenes using a fixed seed and sample benchmark examples only from these held-out scenes. No held-out scene appears in either the dense-depth or instruction-tuning corpora.

\begin{table}[!htbp]
    \centering
    \caption{Compact statistics for the dense and instance-level depth benchmarks. Depth-VQA contains 1,503 questions in total.}
    \label{tab:benchmark_statistics}
    \scriptsize
    \setlength{\tabcolsep}{3.5pt}
    \renewcommand{\arraystretch}{0.96}
    \begin{tabular}{@{}p{0.18\linewidth}p{0.18\linewidth}p{0.17\linewidth}p{0.39\linewidth}@{}}
        \toprule
        Component & Sampling unit & Scale or coverage & Task composition \\
        \midrule
        Level-0 dense depth & Image & 4,228 images from nine datasets & 500 images from each of eight datasets and all 228 Booster images \\
        Level-1 instance depth & Grounded-object query & 757 queries from seven datasets & Center, nearest, farthest, mean, and median metric depth; box or referring prompts \\
        Level-2 Relative & Multi-object question & 370 questions & Spatial Pair (86), Spatial Group (88), and Depth Ordering (196) \\
        Level-2 Metric & Multi-object question & 376 questions & Metric Pair (46 threshold + 48 numerical gap), Spatial Metric Pair (94), Spatial Metric Group (94), and Metric Set Decision (94) \\
        \bottomrule
    \end{tabular}
\end{table}

\subsection{Level-0 Dense-Depth Evaluation}

\paragraph{Sampling and ground truth.} We select 500 frames from each of Argoverse~2, Waymo, nuScenes, DDAD, Taskonomy, ScanNet++, HyperSim, and SUN RGB-D using a fixed random seed, and use all 228 Booster images spanning 38 scenes and six exposure settings. All selected frame identities are removed from the dense training manifest. The automotive datasets are evaluated against their original sensor-derived metric ground truth rather than depth-completion outputs. Booster disparity is converted to metric depth using its official calibration and evaluated with the official validity mask.

\paragraph{Scoring.} Ground-truth camera intrinsics are supplied to every method evaluated in Table~\ref{tab:dense_main}. Predictions are compared directly in metric scale without median scaling or scale-and-shift alignment. For the first eight datasets, valid pixels must be finite and lie within $[0.05,250]$ meters; Booster follows its official mask and $[0.001,10]$ meter range. We compute AbsRel, RMSE, log RMSE, SILog, and $\delta_1$, $\delta_2$, and $\delta_3$ over valid pixels. The main comparison reports $\delta_1$, with the cross-dataset score computed as the unweighted mean of the nine dataset-level results.

\subsection{Depth-VQA Evaluation}
Level~1 metrics are computed over the requested instance-depth values after parsing the numerical answer in meters. For the Level~2 Relative track, a selection answer is correct only if it contains the target identifier, and an ordering answer is correct only if the full predicted identifier sequence matches the reference. Relative Overall Acc is computed over the 86 Spatial Pair, 88 Spatial Group, and 196 Depth Ordering questions ($370$ in total). In the Metric track, Metric Pair separates 46 \path{metric_gap_threshold} questions scored by accuracy from 48 \path{metric_depth_gap} questions scored by $\delta_1$. Spatial Metric Pair reports $\delta_1$ on its 94 \path{spatial_metric_gap} questions. Metric Overall Acc divides the total correct answers on the threshold, Spatial Metric Group, and Metric Set Decision questions by $46+94+94=234$. The 142 numerical gap questions are excluded from Overall Acc and evaluated in their respective $\delta_1$ columns.

\subsection{General-Capability Benchmarks and Baselines}

\paragraph{Benchmarks.} Table~\ref{tab:general_vqa} covers MathVision~\citep{wang2024mathvision}, MathVista~\citep{lu2024mathvista}, MMMU~\citep{yue2024mmmu}, MMMU-Pro~\citep{yue2024mmmupro}, RealWorldQA~\citep{xai2024realworldqa}, MMStar~\citep{chen2024mmstar}, MMBench~\citep{liu2024mmbench}, HallusionBench~\citep{guan2024hallusionbench}, AI2D~\citep{kembhavi2016ai2d}, CC-OCR~\citep{yang2025ccocr}, OCRBench~\citep{liu2023ocrbench}, CharXiv~\citep{wang2024charxiv}, ERQA~\citep{geminirobotics2025erqa}, EmbSpatialBench~\citep{du2024embspatial}, and Ego3D-Bench~\citep{gholami2026ego3d}.

\paragraph{Baselines.} For Level-0 dense prediction, the specialized vision models are Metric3D~\citep{yin2023metric3d}, Metric3D v2~\citep{hu2024metric3dv2}, UniDepth V2~\citep{piccinelli2025unidepthv2}, Depth Anything V1--V3~\citep{yang2024depthanything,yang2024depthanythingv2,lin2025depthanything3}, and MetricAnything~\citep{ma2026metricanything}; the depth-aware VLMs are DepthLM~\citep{cai2026depthlm} and DepthVLM~\citep{yu2026depthvlm}. For instance-level Depth-VQA, the general-purpose VLMs are Qwen3.5-4B~\citep{qwen2026qwen35}, Gemini-3.1-Pro-Preview~\citep{google2026gemini31pro}, Qwen3.7-plus~\citep{qwen2026qwen37plus}, and Seed-2.1-Turbo~\citep{bytedance2026seed21}, while the metric-depth VLMs are DepthLM-12B, DepthVLM-4B, and Youtu-VL-4B-Instruct.

\subsection{Agent-Pipeline Baseline Implementation}
\label{app:agent_pipeline}

The agent baselines in Table~\ref{tab:instance_depth_vqa} use a fixed tool-augmented inference pipeline. For each benchmark question, MetricAnything receives the original RGB image and the camera intrinsics provided by the benchmark and predicts a pixel-level metric depth map in meters. We then crop this prediction with each benchmark-provided grounding box and compute a structured set of regional depth statistics for the corresponding identifier, such as \texttt{obj\_i}. The resulting object-indexed depth table is supplied together with the original image and question to either Qwen3.5-4B or Qwen3.7-plus. The reasoner returns its final response between \texttt{<answer>} and \texttt{</answer>}, and the parsed response is evaluated by the original benchmark scorer. The pipeline therefore exposes predicted metric measurements through an explicit textual table while preserving the benchmark questions, object grounding, and scoring protocol; benchmark boxes are used only to extract regions from the predicted depth map, and no ground-truth depth values are given to the reasoner.

\section{Additional Interface Ablation}

\subsection{Disentangling Interface and Metric Supervision}
\label{app:interface_metric_ablation}

Table~\ref{tab:interface_metric_ablation} tests whether the gain from the dense-to-language interface can be explained by additional supervision on metric values alone. The new \emph{No interface + metric supervision} control removes the interface and adds auxiliary language targets for every instance referenced by a QA example. These targets ask the model to predict the instance's center, mean, median, nearest, and farthest depths (e.g., ``Given $\mathrm{obj}_1$, predict its center depth'') and are optimized with the language loss. The remaining three rows reproduce the corresponding configurations from Table~\ref{tab:interface_ablation}.

\clearpage
\begin{table}[!htbp]
    \centering
    \caption{Disentangling direct metric supervision from the dense-to-language interface on Level-2 Relative and Metric Overall Acc.}
    \label{tab:interface_metric_ablation}
    \footnotesize
    \setlength{\tabcolsep}{4pt}
    \renewcommand{\arraystretch}{0.92}
    \begin{tabular}{lcc}
        \toprule
        Variant & Relative Overall & Metric Overall \\
        \midrule
        No interface & 0.5838 & 0.7692 \\
        No interface + metric supervision & 0.5865 & 0.7949 \\
        Interface + $\mathcal{L}_{\mathrm{ans}}$ & 0.5838 & 0.7735 \\
        Full interface & \textbf{0.6270} & \textbf{0.8504} \\
        \bottomrule
    \end{tabular}
\end{table}

Without the interface, direct metric supervision improves Relative Overall Acc by 0.27 percentage points and Metric Overall Acc by 2.56 points. The larger Metric gain confirms that the auxiliary language targets provide useful numerical depth supervision. However, the full interface remains ahead of this control by 4.05 points on Relative Overall Acc and 5.56 points on Metric Overall Acc. Within this comparison, metric supervision therefore explains only part of the improvement: the full interface provides an additional benefit beyond exposing the language model to referenced-object depth values through auxiliary text targets.

\section{Inference Efficiency}
\label{app:inference_efficiency}

We compare three inference modes with end-of-sequence stopping disabled and a fixed output length of 64 tokens. \emph{No-interface} generates answers without geometry tokens. \emph{Two-pass} first extracts geometry and then reprocesses the multimodal prefix with the geometry tokens to generate the answer. \emph{Single-prefill} reuses the prefix cache and processes the geometry tokens as a continuation, as described in Appendix~\ref{app:cached_geometry_details}. Table~\ref{tab:inference_efficiency} reports generation latency, end-to-end latency, token throughput, and peak allocated GPU memory under this fixed-length setting.

\begin{table}[htbp]
    \centering
    \caption{Inference efficiency with a fixed output length of 64 tokens and end-of-sequence stopping disabled. Single-prefill is the cached geometry-continuation scheme used by DepthEvidence.}
    \label{tab:inference_efficiency}
    \footnotesize
    \setlength{\tabcolsep}{5pt}
    \renewcommand{\arraystretch}{1.10}
    \begin{tabular}{@{}lcccc@{}}
        \toprule
        Mode & \shortstack{Generation\\latency (s)$\downarrow$} & \shortstack{End-to-end\\latency (s)$\downarrow$} & \shortstack{Throughput\\(tokens/s)$\uparrow$} & \shortstack{Peak allocated\\memory (GiB)$\downarrow$} \\
        \midrule
        No-interface & 4.174 & 4.289 & 15.333 & 10.36 \\
        Two-pass & 4.535 & 4.710 & 14.111 & 13.56 \\
        Single-prefill & 4.323 & 4.491 & 14.804 & 13.73 \\
        \bottomrule
    \end{tabular}
\end{table}

Single-prefill reduces generation latency from 4.535 to 4.323~s and end-to-end latency from 4.710 to 4.491~s relative to two-pass inference, reductions of 4.67\% and 4.65\%, respectively. Throughput increases by 4.91\%, from 14.111 to 14.804 tokens/s. Relative to no-interface inference, the end-to-end overhead is 0.202~s (4.71\%). These results support reusing the multimodal prefix to limit the runtime cost of incorporating dense geometry into language reasoning. The benefit is in latency rather than memory: single-prefill uses 13.73~GiB of peak allocated memory, 0.17~GiB more than two-pass and 3.37~GiB more than no-interface. Thus, cached continuation reduces repeated prefix computation while retaining an additional memory cost for geometry-conditioned inference.

\needspace{0.55\textheight}
\section{Additional Qualitative Results}
\label{app:qualitative_results}

\subsection{Dense Metric Depth}
\begin{figure}[!htbp]
    \centering
    \includegraphics[width=0.9 \linewidth]{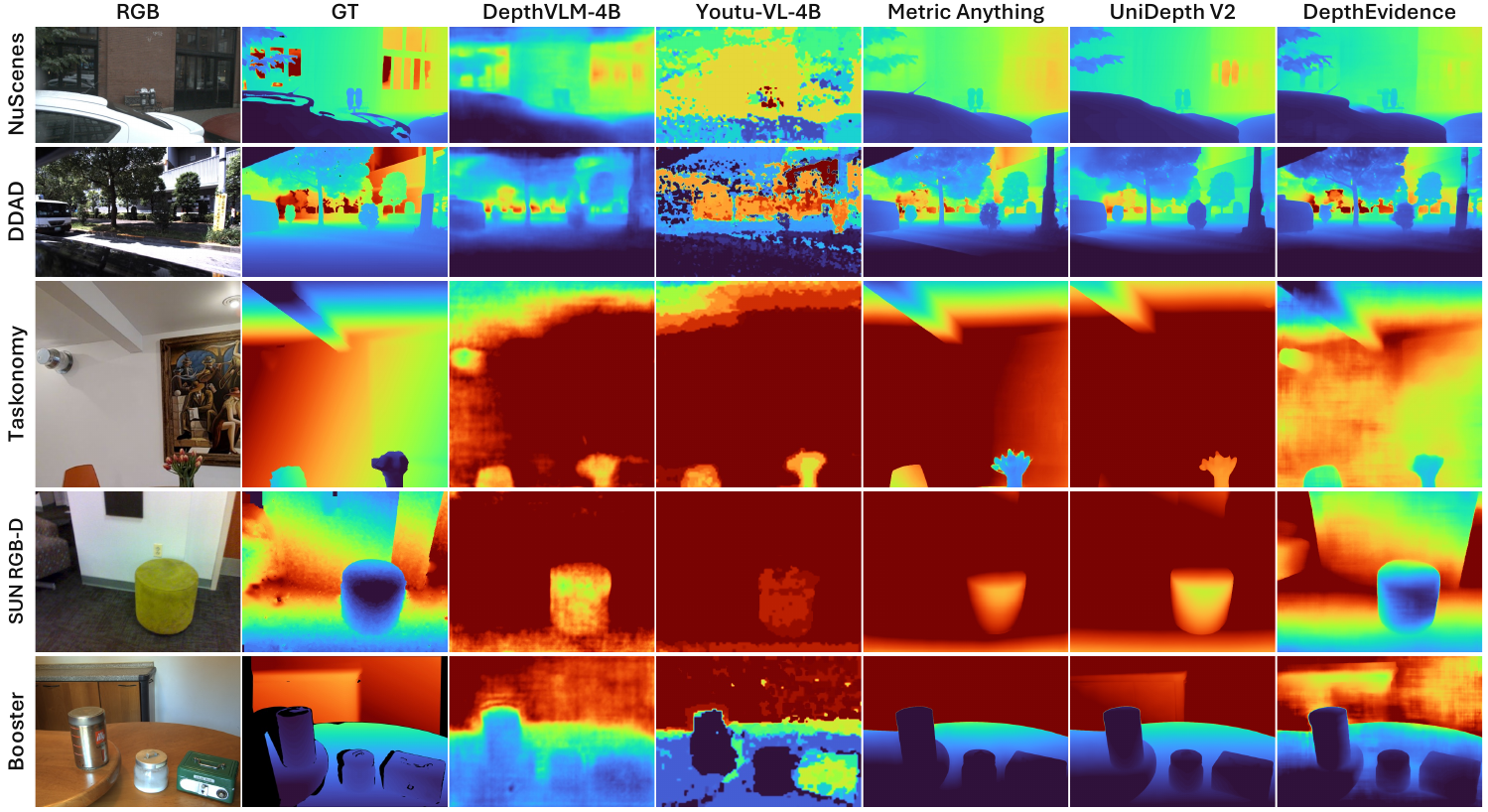}
    \caption{Additional qualitative dense metric depth comparisons on NuScenes, DDAD, Taskonomy, SUN RGB-D, and Booster. Columns show RGB, ground truth (GT), DepthVLM-4B, Youtu-VL-4B-Instruct, MetricAnything, UniDepth V2, and DepthEvidence. Outdoor GT is densely completed for visualization only; quantitative evaluation uses the original sparse sensor GT. All depth maps within a row use the same color scale.}
    \label{fig:dense_metric_qualitative_appendix}
\end{figure}

Figure~\ref{fig:dense_metric_qualitative_appendix} extends the qualitative comparison in Section~\ref{sec:dense_qualitative} to the five remaining dense benchmarks.

\input{sections/10_case_studies_appendix}

%% file: sections/10_case_studies_appendix.tex
\raggedbottom
\newcommand{\casecorrect}{\raisebox{-0.05ex}{\scalebox{1.35}{\textcolor{green!50!black}{\ensuremath{\checkmark}}}}}
\newcommand{\casewrong}{\raisebox{-0.05ex}{\scalebox{1.35}{\textcolor{red!75!black}{\ensuremath{\times}}}}}
\subsection{Instance-Level Depth-VQA}
\label{app:depth_vqa_cases}
We provide selected qualitative cases, including the input image, question, ground-truth answer, and the outputs of Gemini-3.1-Pro, DepthEvidence without the interface, and DepthEvidence. A green checkmark (\casecorrect) denotes a correct benchmark prediction, and a red cross (\casewrong) denotes an incorrect one. Box-grounded questions are paired with the annotated box image, whereas referring-prompt questions retain the original image. Level~1 cases are grouped by dataset; Level~2 cases follow the subbench taxonomy of Table~\ref{tab:level2_benchmark} and cover every Relative Depth and Metric Depth subbench. Examples reproduced in Figure~\ref{fig:interface_cases} are omitted here, except for the sole selected Depth Ordering example.

\newcommand{\depthvqacase}[8]{%
\par\noindent
\begin{minipage}[t]{0.405\linewidth}
\vspace{0pt}\centering
\includegraphics[width=\linewidth,height=0.235\textheight,keepaspectratio]{#3}
\end{minipage}\hfill
\begin{minipage}[t]{0.565\linewidth}
\vspace{0pt}\raggedright
{\scriptsize\textit{#2}}\par
\vspace{0.6mm}{\footnotesize\textbf{Question.} #4\par}
\vspace{0.8mm}
{\footnotesize
\textbf{Ground truth:} #5\par
\textbf{Gemini-3.1-Pro:} #6\par
\textbf{DepthEvidence w/o Interface:} #7\par
\textbf{DepthEvidence:} \textbf{#8}\par}
\end{minipage}
\par\vspace{1.2mm}\hrule\vspace{2.8mm}
}

\subsubsection{Level-1 Cases Grouped by Dataset}
These examples span center-point, mean, median, nearest, and farthest instance-depth queries. The displayed target specification is the semantic reference seen by the model; box coordinates remain in the question for box-grounded cases.

\medskip\noindent\textbf{LingBot-Depth}\par\smallskip

\depthvqacase
{Case 17}
{Mean depth; box prompt}
{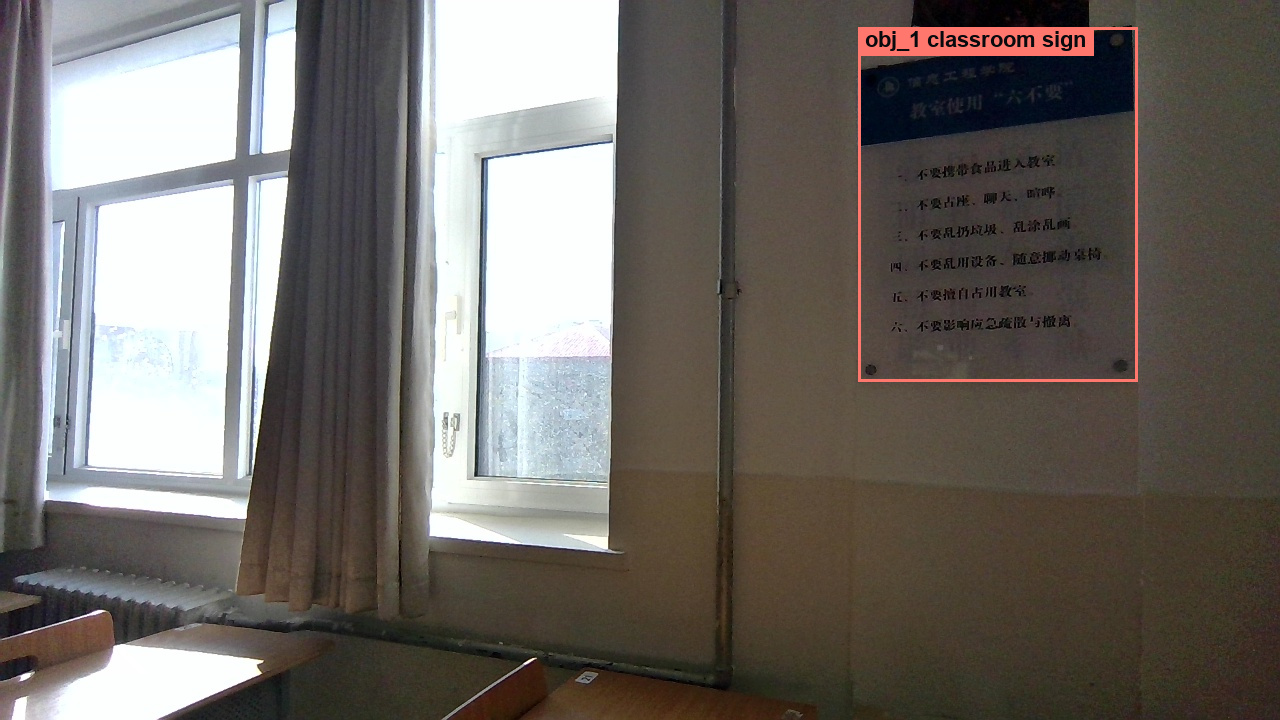}
{Given this image and the target object, estimate the average depth of the target region from the camera. Target: classroom sign with bbox\_2d [670, 38, 888, 529]}
{1.25 m}
{2.11 m\nobreak\hspace{0.25em}\casewrong}
{1.31 m\nobreak\hspace{0.25em}\casecorrect}
{1.32 m\nobreak\hspace{0.25em}\casecorrect}

\medskip\noindent\textbf{Waymo}\par\smallskip


\depthvqacase
{Case 39}
{Farthest depth; box prompt}
{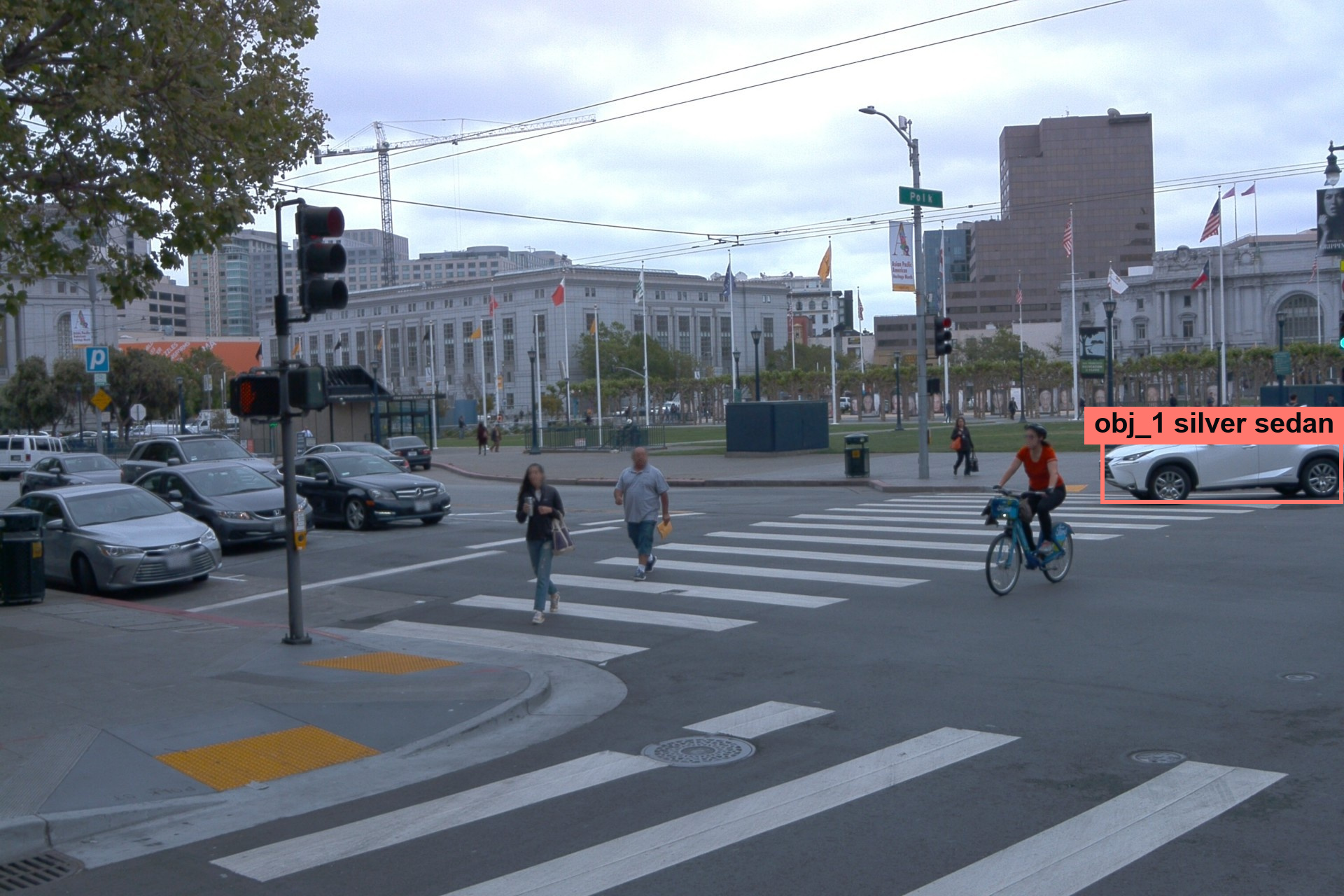}
{Given this image and the target object, estimate the farthest depth of the target region from the camera. Target: silver sedan with bbox\_2d [819, 454, 999, 562]}
{27.167 m}
{41.11 m\nobreak\hspace{0.25em}\casewrong}
{26.62 m\nobreak\hspace{0.25em}\casecorrect}
{27.3 m\nobreak\hspace{0.25em}\casecorrect}

\depthvqacase
{Case 43}
{Median depth; referring prompt}
{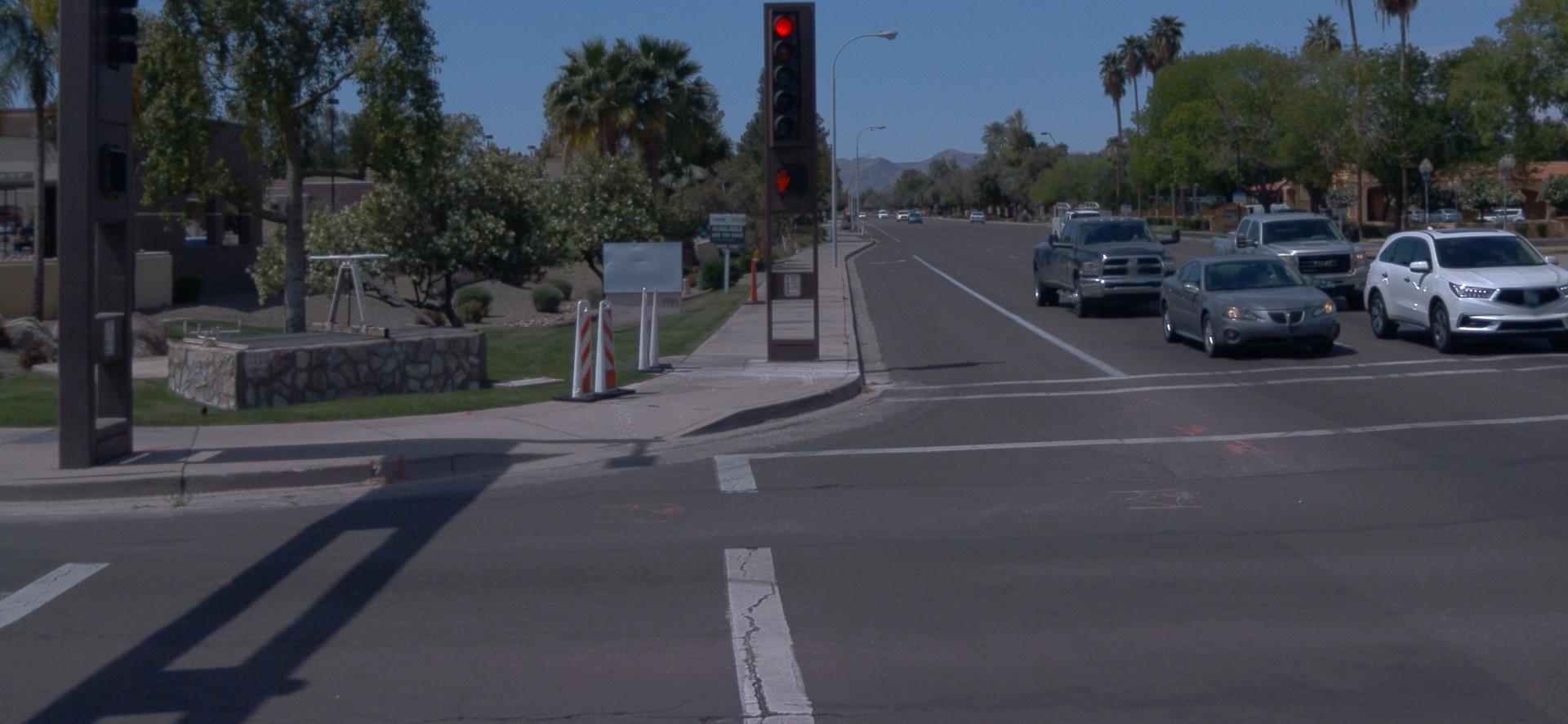}
{Given this image and the target object, estimate the median depth of the target region from the camera. Target: pole referred as "the tall, dark metal pole on the far left sidewalk near the traffic light"}
{16 m}
{10.11 m\nobreak\hspace{0.25em}\casewrong}
{15 m\nobreak\hspace{0.25em}\casecorrect}
{15.56 m\nobreak\hspace{0.25em}\casecorrect}


\medskip\noindent\textbf{DDAD}\par\smallskip

\depthvqacase
{Case 668}
{Mean depth; box prompt}
{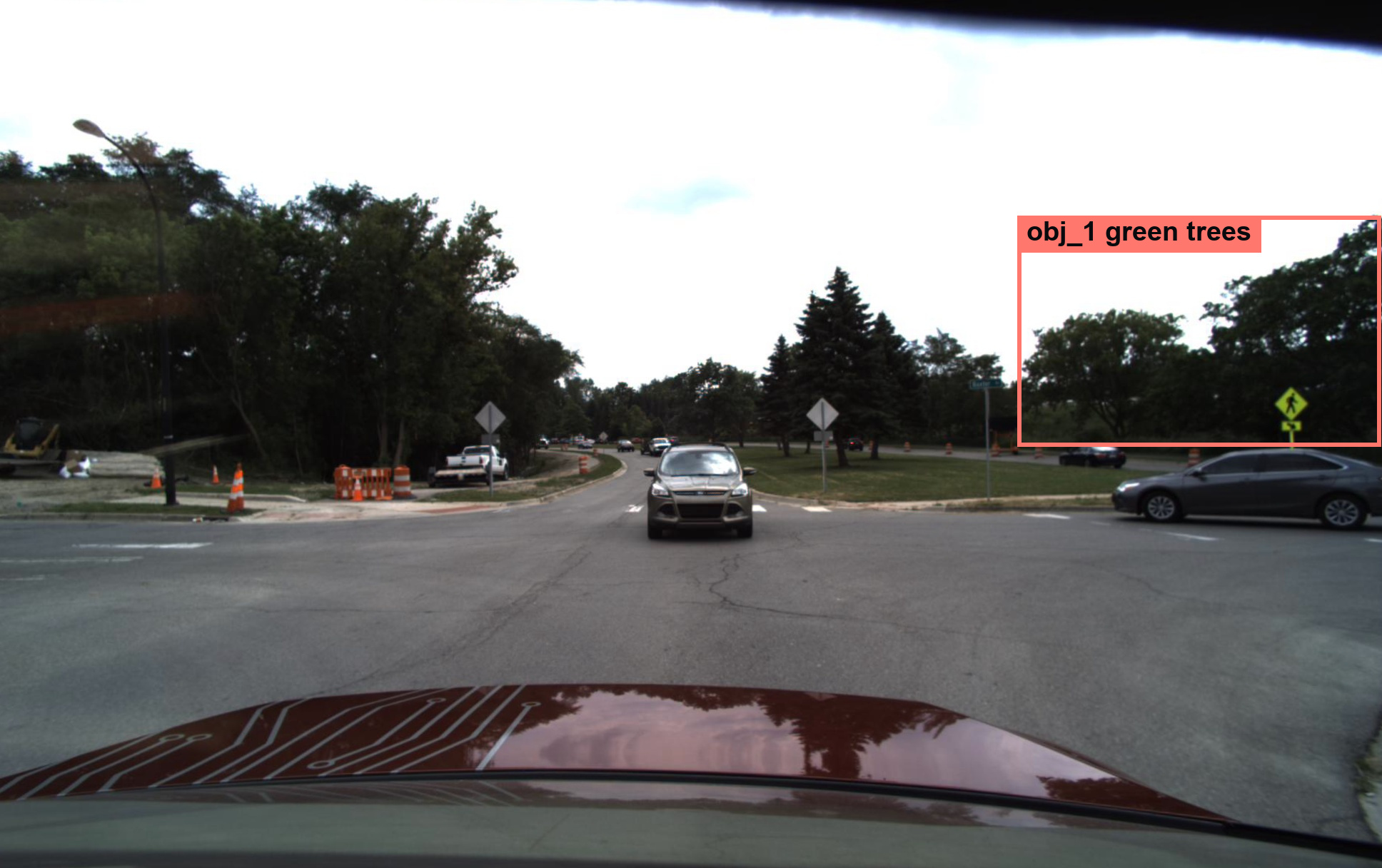}
{Given this image and the target object, estimate the average depth of the target region from the camera. Target: green trees with bbox\_2d [736, 248, 999, 514]}
{50.94 m}
{101.12 m\nobreak\hspace{0.25em}\casewrong}
{52.4 m\nobreak\hspace{0.25em}\casecorrect}
{55.3 m\nobreak\hspace{0.25em}\casecorrect}

\medskip\noindent\textbf{Argoverse 2}\par\smallskip


\depthvqacase
{Case 696}
{Nearest depth; referring prompt}
{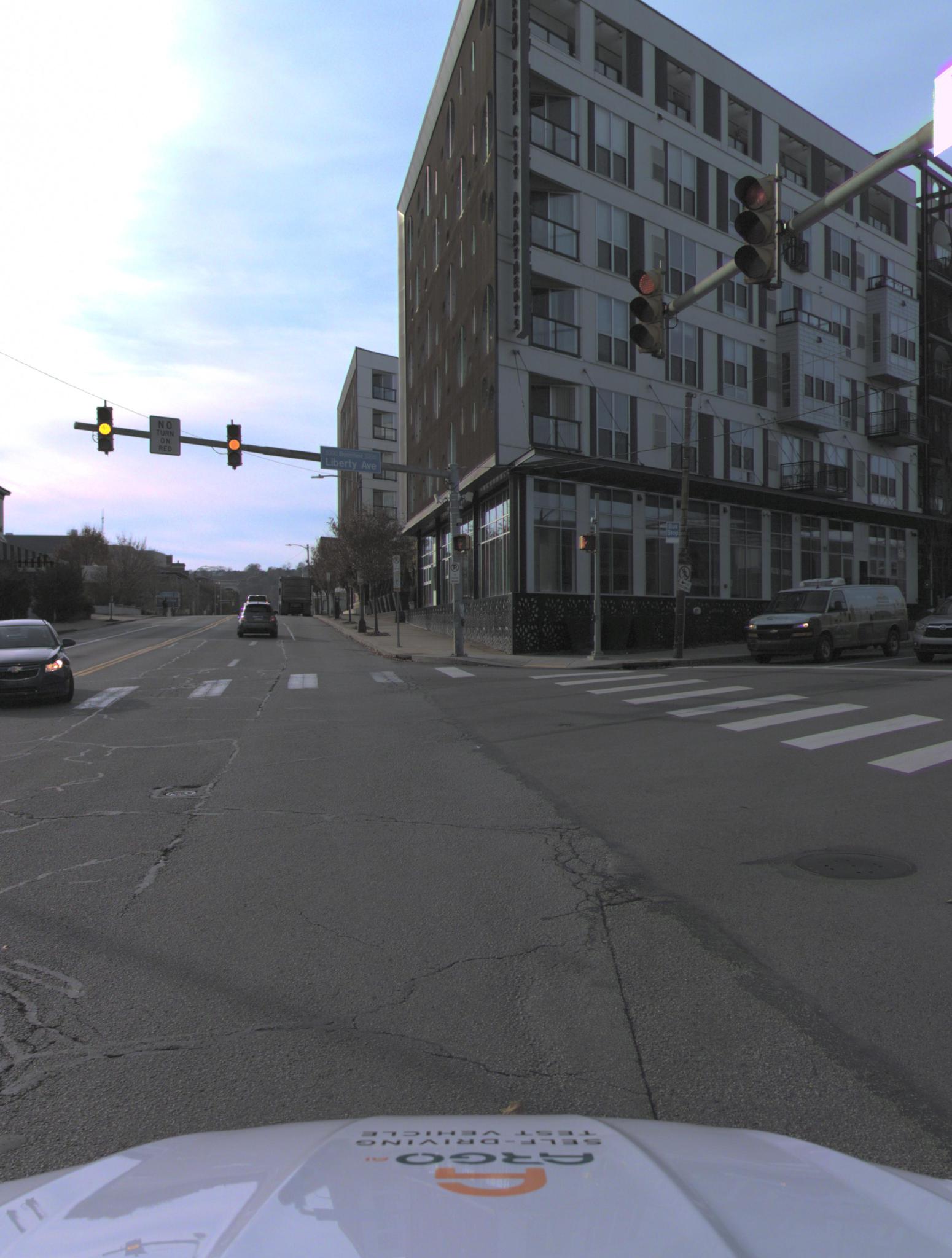}
{Given this image and the target object, estimate the nearest depth of the target region from the camera. Target: white car hood referred as "the white car hood with 'ARGO SELF-DRIVING TEST VEHICLE' text at the bottom of the image"}
{1.15 m}
{1.8 m\nobreak\hspace{0.25em}\casewrong}
{1.06 m\nobreak\hspace{0.25em}\casecorrect}
{1.03 m\nobreak\hspace{0.25em}\casecorrect}

\depthvqacase
{Case 702}
{Nearest depth; box prompt}
{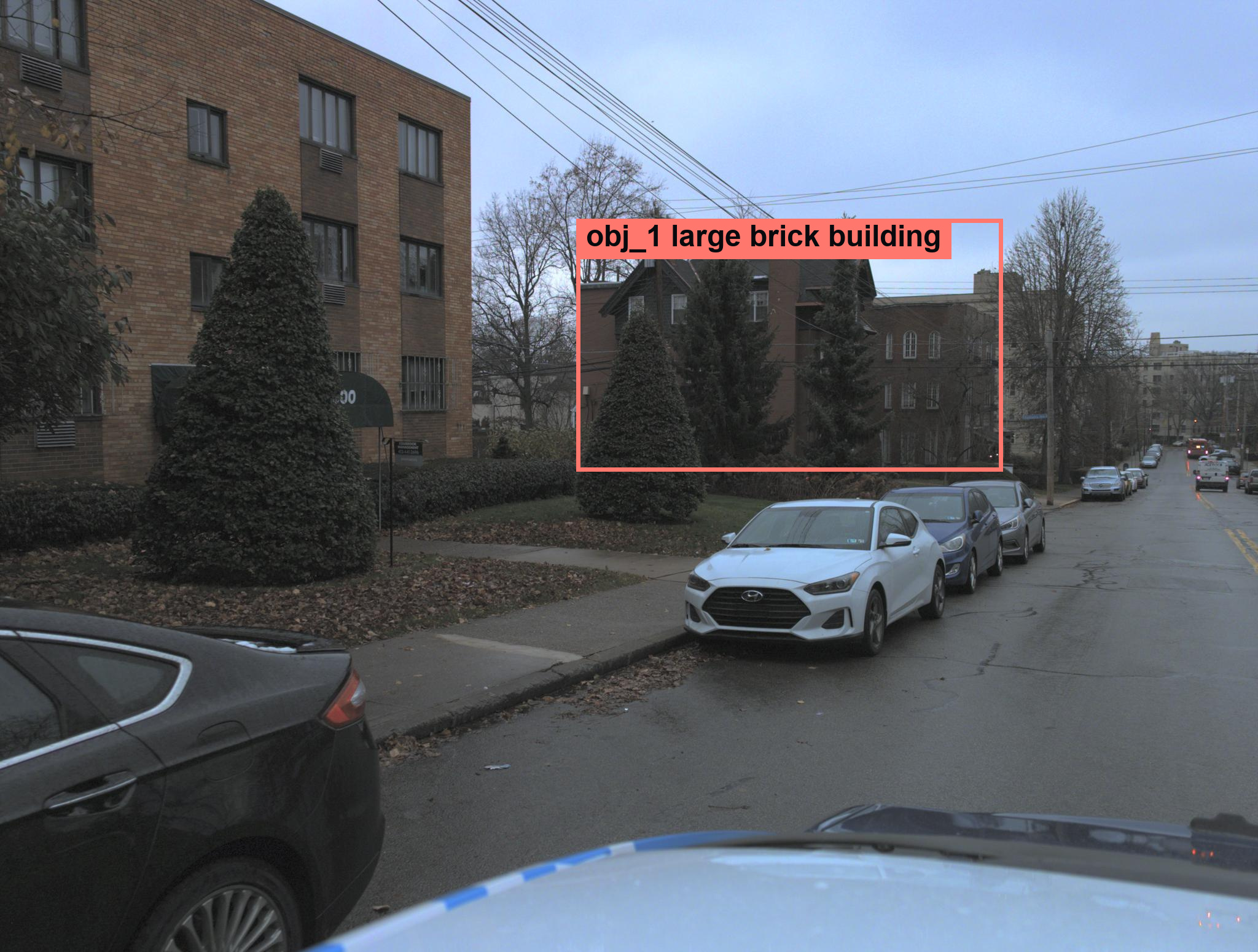}
{Given this image and the target object, estimate the nearest depth of the target region from the camera. Target: large brick building with bbox\_2d [458, 230, 797, 495]}
{37.46 m}
{10.82 m\nobreak\hspace{0.25em}\casewrong}
{40.5 m\nobreak\hspace{0.25em}\casecorrect}
{36.5 m\nobreak\hspace{0.25em}\casecorrect}

\medskip\noindent\textbf{nuScenes}\par\smallskip

\depthvqacase
{Case 718}
{Median depth; box prompt}
{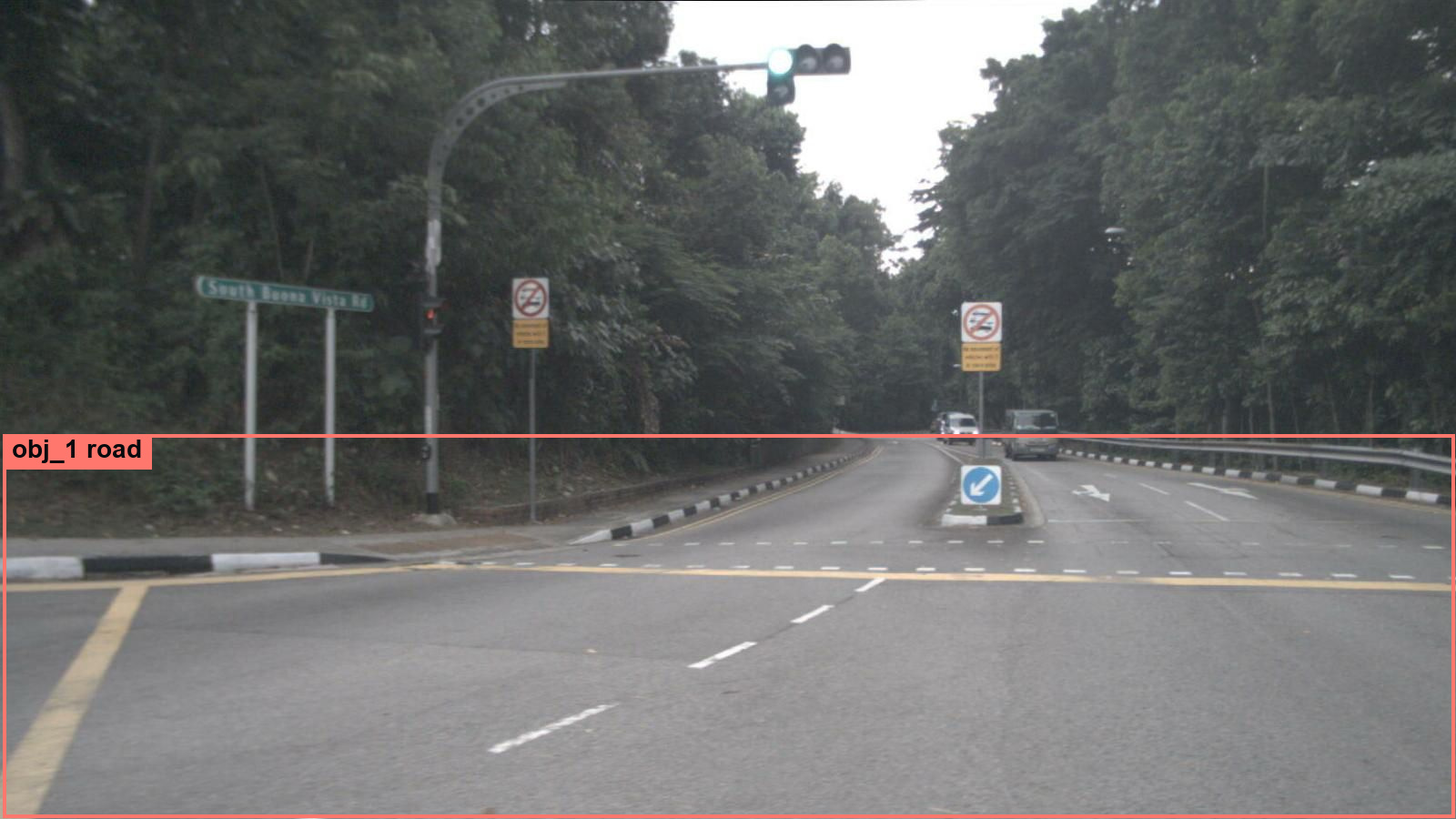}
{Given this image and the target object, estimate the median depth of the target region from the camera. Target: road with bbox\_2d [2, 530, 999, 998]}
{7.19 m}
{14.11 m\nobreak\hspace{0.25em}\casewrong}
{7.53 m\nobreak\hspace{0.25em}\casecorrect}
{7.06 m\nobreak\hspace{0.25em}\casecorrect}

\medskip\noindent\textbf{Taskonomy}\par\smallskip

\depthvqacase
{Case 721}
{Mean depth; box prompt}
{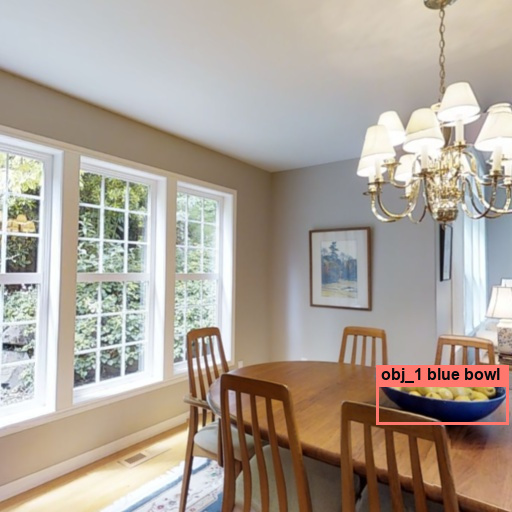}
{Given this image and the target object, estimate the average depth of the target region from the camera. Target: blue bowl with bbox\_2d [734, 713, 992, 828]}
{1.53 m}
{1.81 m\nobreak\hspace{0.25em}\casecorrect}
{1.4 m\nobreak\hspace{0.25em}\casecorrect}
{1.48 m\nobreak\hspace{0.25em}\casecorrect}

\subsubsection{Level-2 Cases Grouped by Table~\ref{tab:level2_benchmark} Subbench}
The cases below retain the same Relative Depth and Metric Depth organization used by the benchmark. Annotated boxes identify the candidate objects and their stable \texttt{obj\_i} identifiers.

\medskip\noindent\textbf{Relative Depth}\par\smallskip

\Needspace{0.30\textheight}
\vspace{1mm}\noindent\textbf{Spatial Pair.}\par\smallskip

\depthvqacase
{Case 841}
{\texttt{spatial\_depth}; box prompt}
{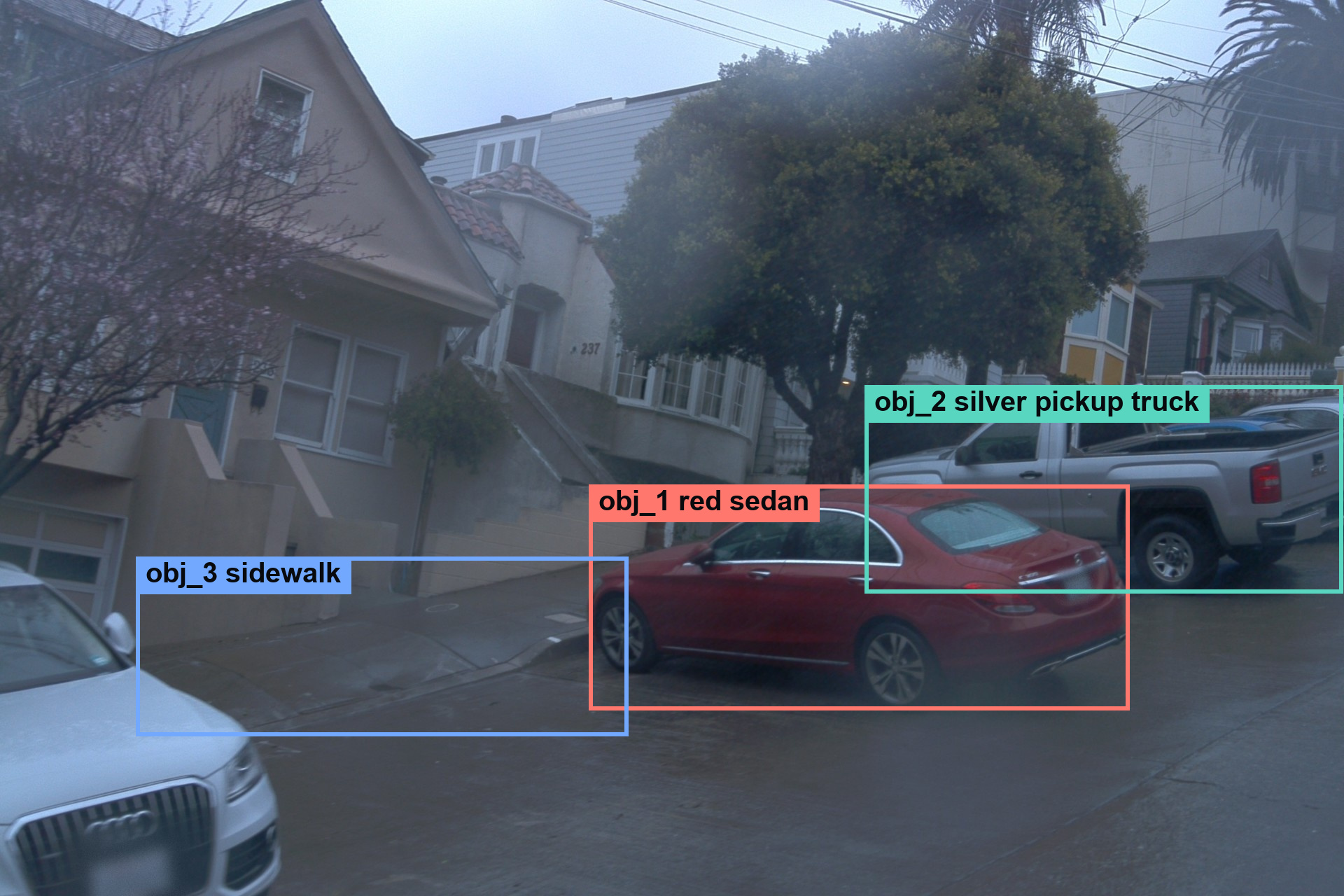}
{Which is closer to the camera, the object to the left of silver pickup truck obj\_2 or sidewalk obj\_3?}
{red sedan obj\_1.}
{obj\_3\nobreak\hspace{0.25em}\casewrong}
{red sedan obj\_1.\nobreak\hspace{0.25em}\casecorrect}
{red sedan obj\_1.\nobreak\hspace{0.25em}\casecorrect}

\Needspace{0.30\textheight}
\vspace{1mm}\noindent\textbf{Spatial Group.}\par\smallskip

\depthvqacase
{Case 788}
{\texttt{spatial\_depth\_group}; box prompt}
{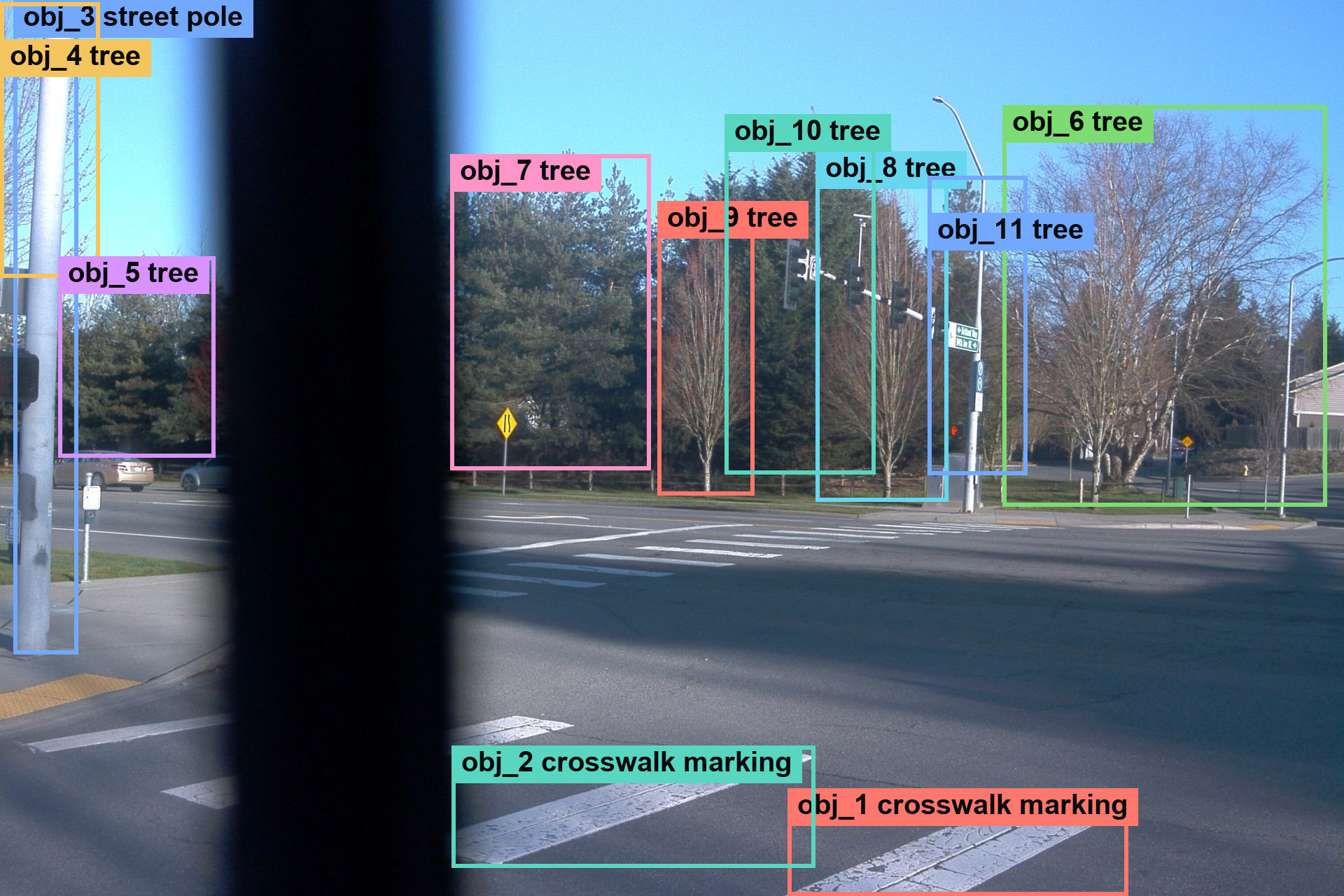}
{Among the objects to the left of tree\_3 obj\_6, which is closest to the camera?}
{crosswalk marking\_1 obj\_1.}
{obj\_3\nobreak\hspace{0.25em}\casewrong}
{crosswalk marking\_1 obj\_1.\nobreak\hspace{0.25em}\casecorrect}
{crosswalk marking\_1 obj\_1.\nobreak\hspace{0.25em}\casecorrect}

\Needspace{0.30\textheight}
\vspace{1mm}\noindent\textbf{Depth Ordering.}\par\smallskip

\depthvqacase
{Case 899}
{\texttt{depth\_ordering\_all}; box prompt}
{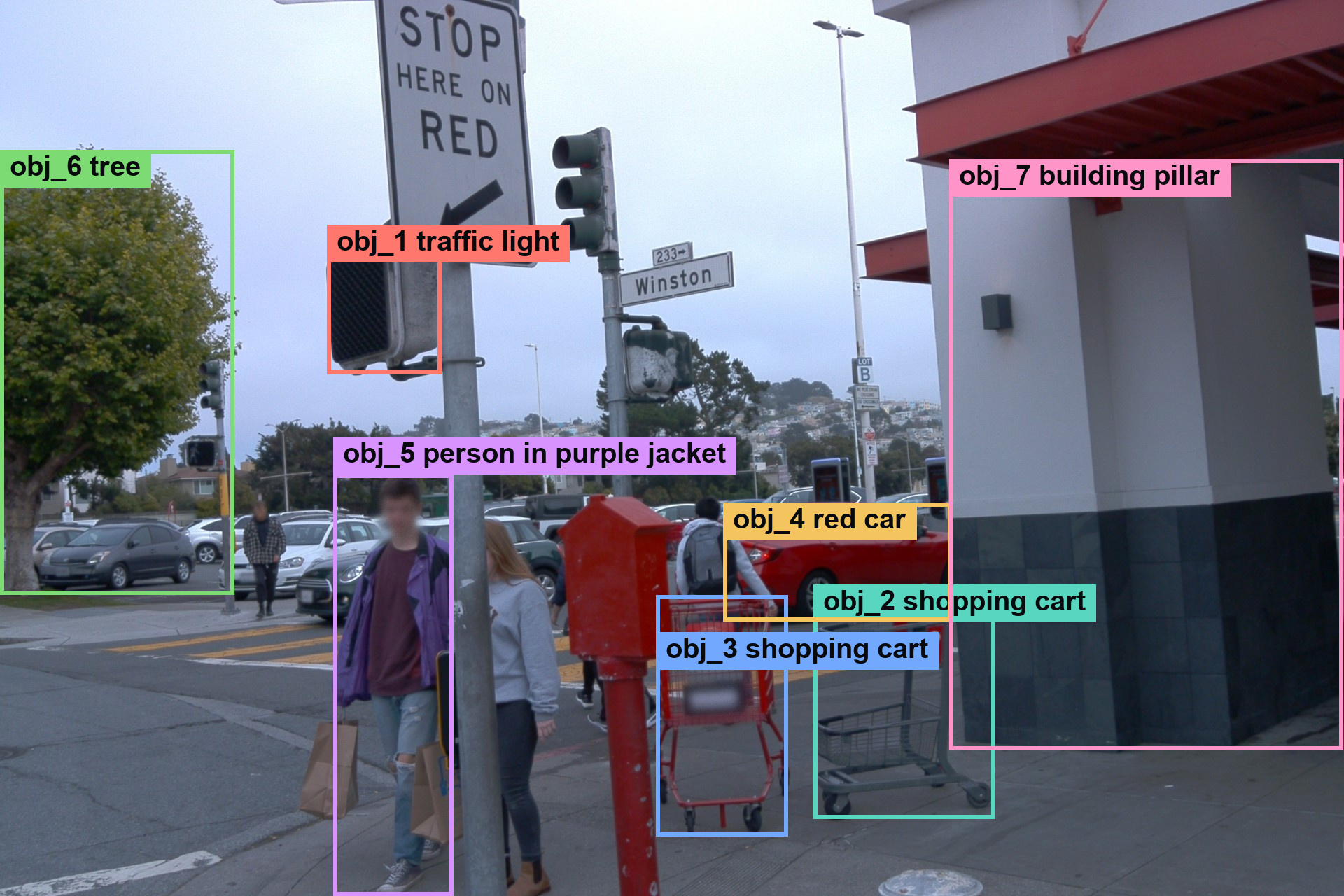}
{Rank the main objects in this scene by their distance to the camera, starting from the closest.}
{traffic light obj\_1, person in purple jacket obj\_5, shopping cart\_2 obj\_3, shopping cart\_1 obj\_2, building pillar obj\_7, red car obj\_4, tree obj\_6.}
{obj\_1, obj\_5, obj\_7, obj\_3, obj\_2, obj\_4, obj\_6\nobreak\hspace{0.25em}\casewrong}
{traffic light obj\_1, person in purple jacket obj\_5, shopping cart\_1 obj\_2, shopping cart\_2 obj\_3, building pillar obj\_7, red car obj\_4, tree obj\_6.\nobreak\hspace{0.25em}\casewrong}
{traffic light obj\_1, person in purple jacket obj\_5, shopping cart\_2 obj\_3, shopping cart\_1 obj\_2, building pillar obj\_7, red car obj\_4, tree obj\_6.\nobreak\hspace{0.25em}\casecorrect}

\medskip\noindent\textbf{Metric Depth}\par\smallskip

\Needspace{0.30\textheight}
\vspace{1mm}\noindent\textbf{Metric Pair.}\par\smallskip

\depthvqacase
{Case 989}
{\texttt{metric\_gap\_threshold}; box prompt}
{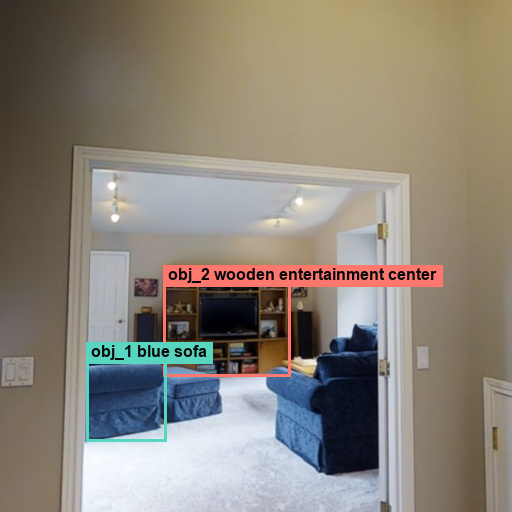}
{Is the average camera depth of wooden entertainment center obj\_2 at least 6.60 m greater than that of blue sofa obj\_1?}
{No.}
{yes. \nobreak\hspace{0.25em}\casewrong}
{no.\nobreak\hspace{0.25em}\casecorrect}
{No.\nobreak\hspace{0.25em}\casecorrect}

\Needspace{0.30\textheight}
\vspace{1mm}\noindent\textbf{Spatial Metric Pair.}\par\smallskip

\depthvqacase
{Case 1005}
{\texttt{spatial\_metric\_gap}; box prompt}
{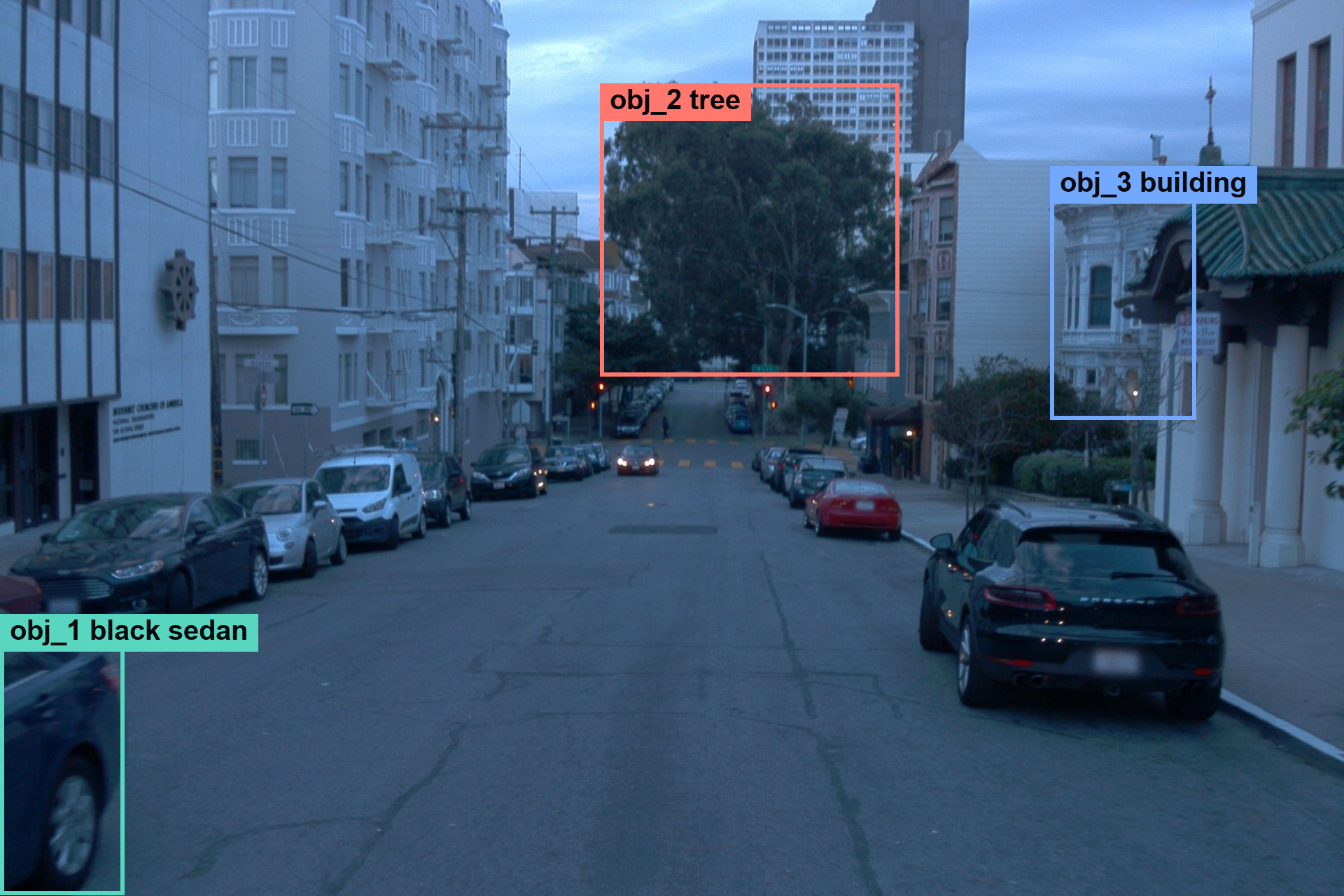}
{What is the absolute difference in average camera depth between the object to the right of black sedan obj\_1 and building obj\_3?}
{38.39 m}
{10.5\nobreak\hspace{0.25em}\casewrong}
{27.00\nobreak\hspace{0.25em}\casewrong}
{40.00 m\nobreak\hspace{0.25em}\casecorrect}

\Needspace{0.30\textheight}
\vspace{1mm}\noindent\textbf{Spatial Metric Group.}\par\smallskip

\depthvqacase
{Case 1063}
{\texttt{spatial\_range\_filter}; box prompt}
{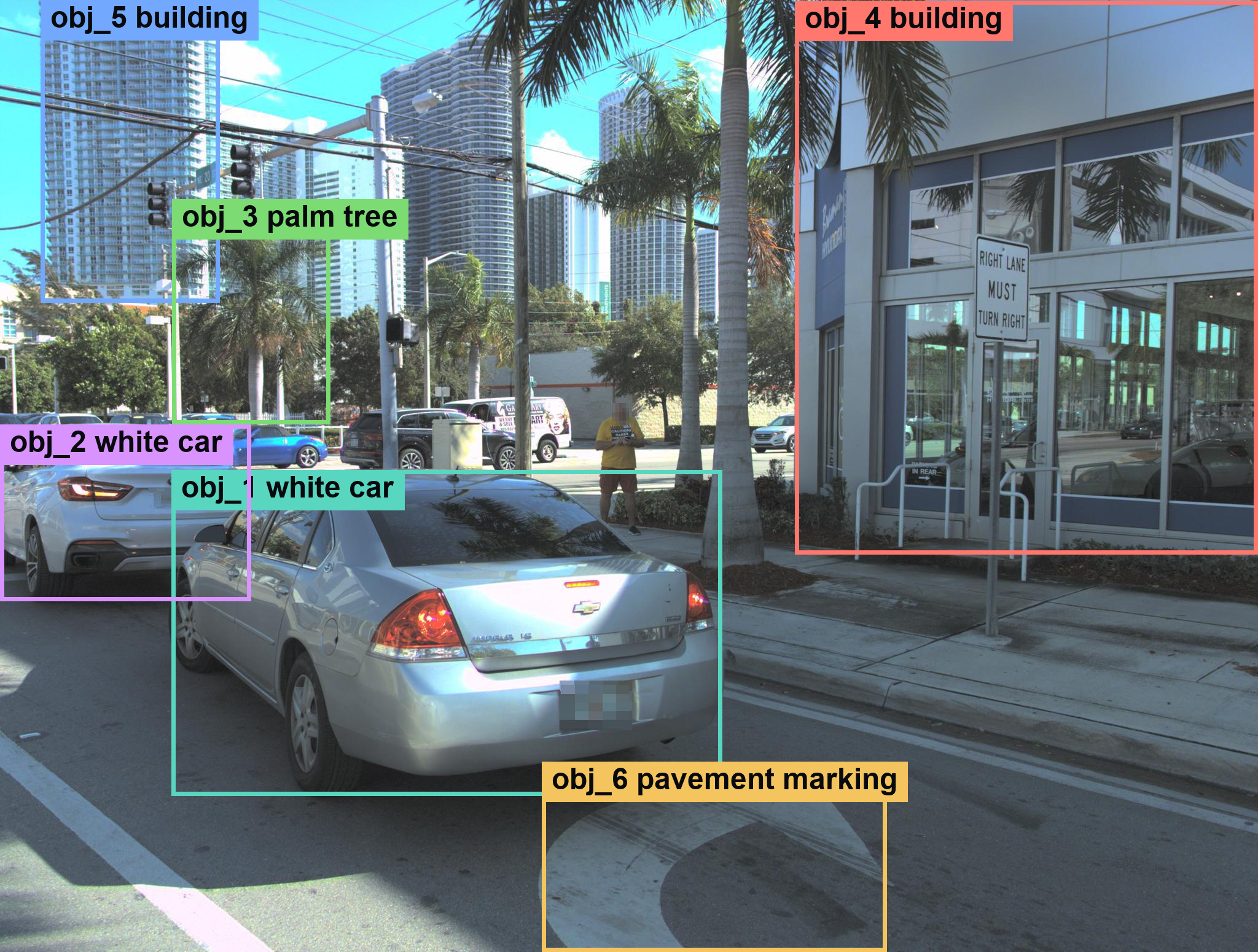}
{Among the referenced objects to the left of building obj\_4, which have an average camera depth between 55.21 m and 79.57 m, inclusive?}
{building obj\_5.}
{obj\_3\nobreak\hspace{0.25em}\casewrong}
{building obj\_5.\nobreak\hspace{0.25em}\casecorrect}
{building obj\_5.\nobreak\hspace{0.25em}\casecorrect}

\depthvqacase
{Case 1070}
{\texttt{compositional\_metric\_constraint}; box prompt}
{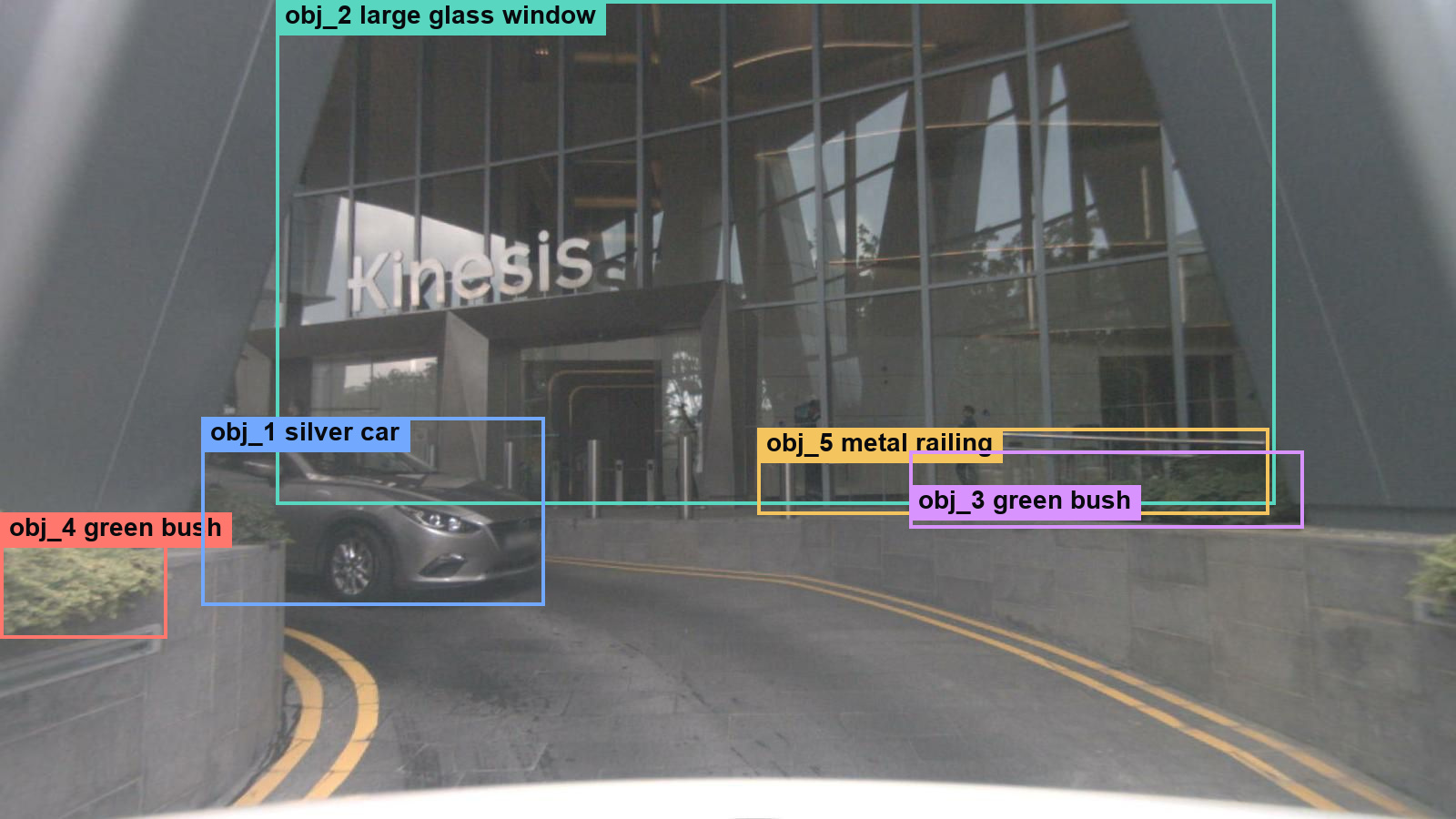}
{Among the referenced objects to the right of green bush obj\_4, select those with an average camera depth greater than 4.50 m but smaller than that of large glass window obj\_2.}
{silver car obj\_1, metal railing obj\_5, green bush obj\_3.}
{obj\_3, obj\_5\nobreak\hspace{0.25em}\casewrong}
{silver car obj\_1, green bush obj\_3, metal railing obj\_5.\nobreak\hspace{0.25em}\casecorrect}
{silver car obj\_1, metal railing obj\_5, green bush obj\_3.\nobreak\hspace{0.25em}\casecorrect}

\Needspace{0.30\textheight}
\vspace{1mm}\noindent\textbf{Metric Set Decision.}\par\smallskip

\depthvqacase
{Case 1093}
{\texttt{metric\_margin\_decision}; box prompt}
{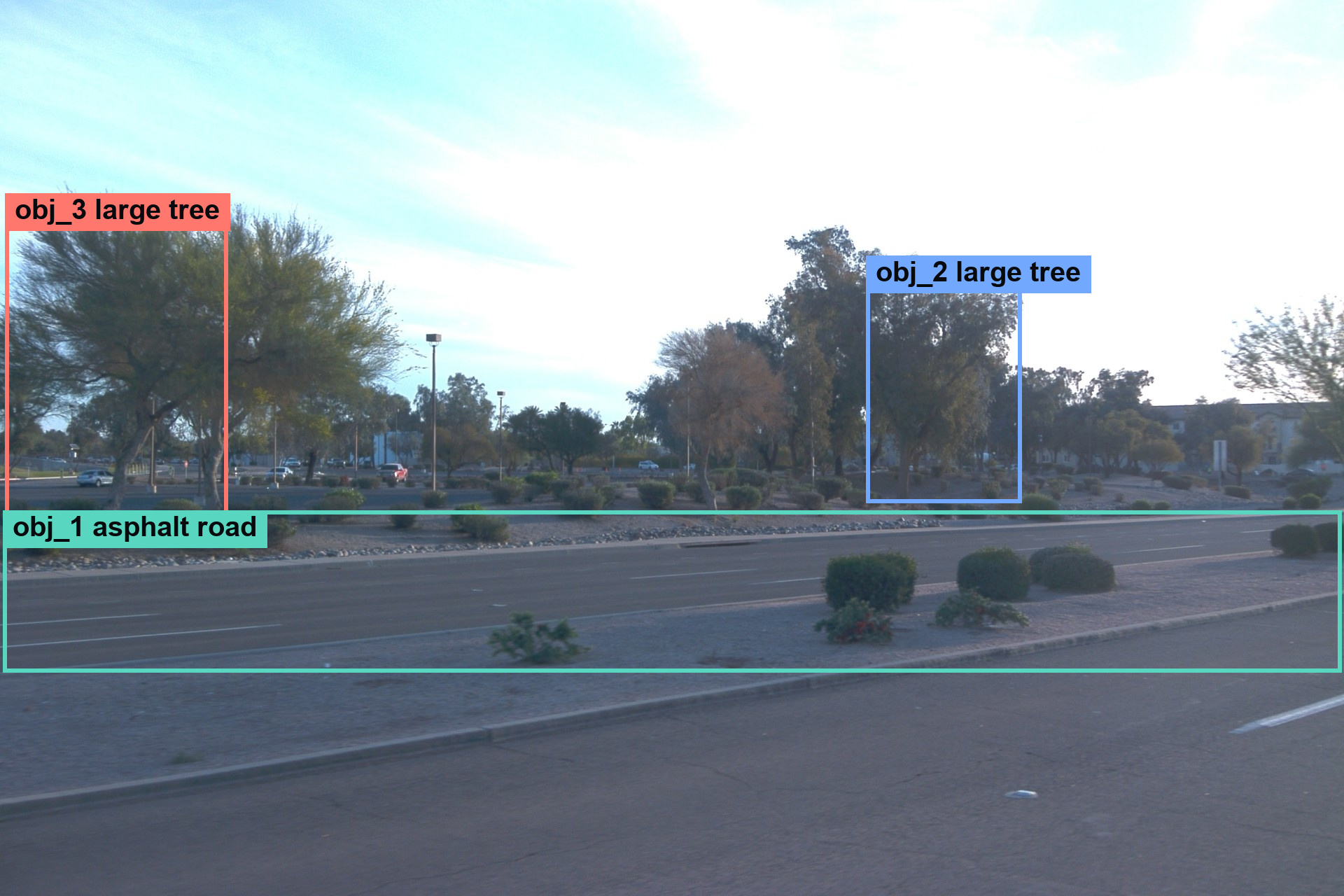}
{Select the closest referenced object only if it is at least 21.00 m closer in average camera depth than every other candidate; otherwise answer ambiguous.}
{ambiguous.}
{obj\_1\nobreak\hspace{0.25em}\casewrong}
{asphalt road obj\_1.\nobreak\hspace{0.25em}\casewrong}
{ambiguous.\nobreak\hspace{0.25em}\casecorrect}

\depthvqacase
{Case 1126}
{\texttt{camera\_safety\_band}; box prompt}
{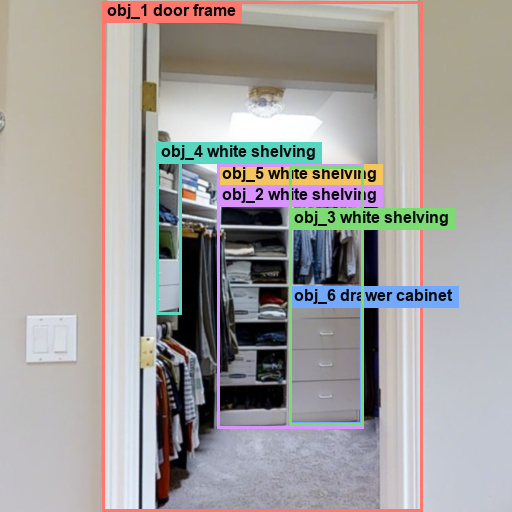}
{Is any referenced object at an average camera depth of no more than 0.80 m?}
{No.}
{yes\nobreak\hspace{0.25em}\casewrong}
{no\nobreak\hspace{0.25em}\casecorrect}
{No.\nobreak\hspace{0.25em}\casecorrect}